\documentclass{article}
\usepackage[nonatbib,preprint]{neurips_2023}
\usepackage[utf8]{inputenc} % allow utf-8 input
\usepackage[T1]{fontenc}    % use 8-bit T1 fonts
\usepackage{hyperref}       % hyperlinks
\usepackage{url}            % simple URL typesetting
\usepackage{booktabs}       % professional-quality tables
\usepackage{amsfonts}       % blackboard math symbols
\usepackage{nicefrac}       % compact symbols for 1/2, etc.
\usepackage{microtype}      % microtypography
\usepackage{xspace}
\usepackage{adjustbox}
\usepackage{graphicx} 
\usepackage{multirow}
\usepackage{MnSymbol}
\newcommand{\name}{X-Detect\xspace}
\usepackage[table,xcdraw]{xcolor}
\usepackage{caption}
\usepackage{subcaption}
\definecolor{Gray}{gray}{0.9}
\usepackage{xurl}

\title{\name: Explainable Adversarial Patch Detection for Object Detectors in Retail}

\author{%
  Omer Hofman\thanks{Corresponding author}\textsuperscript{\rm 1}
    Amit Giloni\textsuperscript{\rm 1},
    Yarin Hayun\textsuperscript{\rm 1},
    Ikuya Morikawa\textsuperscript{\rm 2},
    Toshiya Shimizu\textsuperscript{\rm 2},\\
    \textbf{Yuval Elovici\textsuperscript{\rm 1}},
    \textbf{Asaf Shabtai\textsuperscript{\rm 1}}\\
    $^1$Ben-Gurion University of the Negev, \quad $^2$Fujitsu Limited \\
\texttt{\{omerhof\thanks{Corresponding author},hacmona,yarinbo\}@post.bgu.ac.il}\\
\texttt{\{morikawa.ikuya,shimizu.toshiya\}@fujitsu.com}\\
\texttt{\{elovici,shabtaia\}@bgu.ac.il}
}

\begin{document}

\maketitle

\begin{abstract}
    Object detection models, which are widely used in various domains (such as retail), have been shown to be vulnerable to adversarial attacks.
    Existing methods for detecting adversarial attacks on object detectors have had difficulty detecting new real-life attacks.
    We present \name, a novel adversarial patch detector that can: \textit{i)} detect adversarial samples in real time, allowing the defender to take preventive action; \textit{ii)} provide explanations for the alerts raised to support the defender's decision-making process, and \textit{iii)} handle unfamiliar threats in the form of new attacks.
    Given a new scene, \name uses an ensemble of explainable-by-design detectors that utilize object extraction, scene manipulation, and feature transformation techniques to determine whether an alert needs to be raised.
    \name was evaluated in both the physical and digital space using five different attack scenarios (including adaptive attacks) and the COCO dataset and our new Superstore dataset.
    The physical evaluation was performed using a smart shopping cart setup in real-world settings and included 17 adversarial patch attacks recorded in 1,700 adversarial videos.
    The results showed that \name outperforms the state-of-the-art methods in distinguishing between benign and adversarial scenes for all attack scenarios while maintaining a 0\% FPR (no false alarms) and providing actionable explanations for the alerts raised. A demo is available.
\end{abstract}
\section{\label{sec:intro}Introduction}
Object detection (OD) models are commonly used in computer vision in various industries, including manufacturing~\cite{song2021object}, autonomous driving~\cite{khatab2021vulnerable}, security surveillance~\cite{kalli2021effective}, and retail~\cite{melek2017survey,fuchs2019towards,cheng2021fashion,cai2021rethinking}.
Since existing retail automated checkout solutions often necessitate extensive changes to a store's facilities (as in Amazon's Just Walk-Out~\cite{green2021super}), a simpler solution based on a removable plugin placed on the shopping cart and an OD model was proposed (both in the literature~\cite{oh2020implementation,santra2019comprehensive} and by industry).\footnote{Shopic: shopic.co}
Since adversarial attacks can compromise an OD model's integrity~\cite{liu2018dpatch,thys2019fooling,zolfi2021translucent,hu2021naturalistic}, such smart shopping carts can also be at risk.
For example, an adversarial customer could place an adversarial patch on a high-cost product (such as an expensive bottle of wine), which would cause the OD model to misclassify it as a cheaper product (such as a carton of milk).
Due to the expansion in retail thefts~\cite{nrss2022,FSR2022}, such attacks would result in a loss of revenue for a retail chain and compromise the solution's trustworthiness, if not addressed.\footnote{Demo: drive.google.com/file/d/1KMRHsqjRucVo0I2HuCkr2cCsu6JWZFwh/ view?usp=sharing}

\begin{figure}[t]
\includegraphics[width=1.0\linewidth]{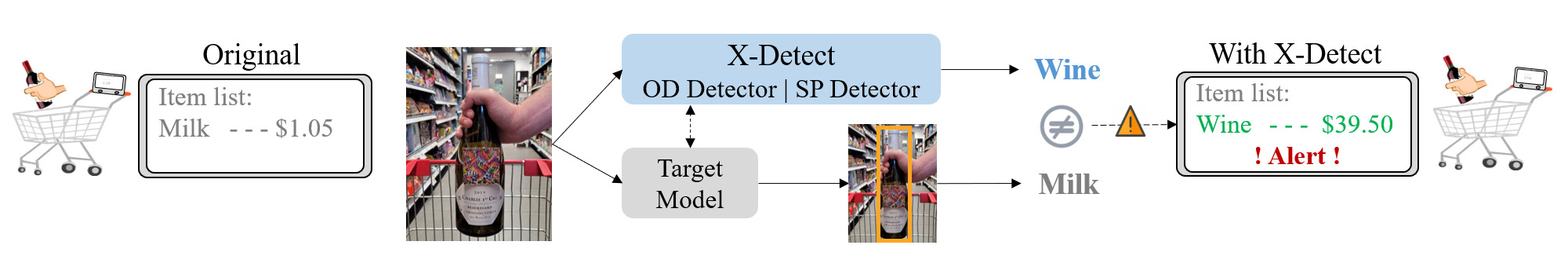}
\centering
\caption{Adversarial theft detection when using \name.}
\label{X-detect_pipline}
\end{figure}

The detection of such physical patch attacks in the retail setting is a challenging task since:
\textit{i)} the defense mechanism (adversarial detector) should raise an alert in an actionable time frame to prevent the theft from taking place;
\textit{ii)} the adversarial detector should provide explanations when an adversarial alert is raised to prevent a retail chain from falsely accusing a customer of being a thief~\cite{ASPD2022};
and \textit{iii)} the adversarial detector should be capable of detecting unfamiliar threats (i.e., adversarial patches in different shapes, colors, and  textures)~\cite{zhang2019limitations}.
Several methods for detecting adversarial patches for OD models have been proposed~\cite{chiang2021adversarial,ji2021adversarial,liu2022segment}, however, none of them provide an adequate solution that addresses all of the above requirements.

We present \name, a novel adversarial detector for OD models which is suitable for real-life settings.
Given a new scene, \name identifies whether the scene contains an adversarial patch, i.e., whether an alert needs to be raised (an illustration of its use is presented in Figure~\ref{X-detect_pipline}).
\name consists of two base-detectors: the \textit{object extraction detector} (OED) and the \textit{scene processing detector} (SPD), each of which alters the attacker's assumed attack environment in order to detect the presence of an adversarial patch. 
The OED changes the attacker's assumed machine learning (ML) task, i.e., object detection, to image classification by utilizing an object extraction model and a customized k-nearest neighbors (KNN) classifier.
The SPD changes the attacker's assumed OD pipeline by adding a scene preprocessing step to limit the effect of the adversarial patch. 
The two base detectors can be used as an ensemble or on their own.

We empirically evaluated \name in both the digital and physical space using five different attack scenarios (including adaptive attacks) that varied in terms of the attacker's level of knowledge.
The evaluation was performed using a variety of OD algorithms, including Faster R-CNN~\cite{ren2015faster}, YOLO~\cite{redmon2018yolov3}, Cascade R-CNN~\cite{cai2019cascade}, and Grid R-CNN.
In the digital evaluation, we digitally placed adversarial patches on objects in the Common Objects in Context (COCO) dataset, which is a benchmark dataset in the OD domain.
In the physical evaluation, we physically placed 17 adversarial patches on objects (products found in retail stores) and created more than 1,700 adversarial videos that were recorded in a real smart shopping cart setup.
For this evaluation, we created the Superstore dataset, which is an OD dataset tailored to the retail domain.
Our evaluation results show that \name can successfully identify digital and physical adversarial patches, outperform state-of-the-art methods (Segment \& Complete and Ad-YOLO) without interfering with the detection in benign scenes (scenes without an adversarial patch), and provide explanations for its output, without being exposed to adversarial examples.
The main contributions of this paper are as follows:
\begin{itemize}
\item To the best of our knowledge, \name is the first adversarial detector capable of providing explainable adversarial detection for OD models; moreover, \name can be employed in any user-oriented domain where explanations are needed, and specifically in retail.

\item \name is a model-agnostic solution.
By requiring only black-box access to the target model, it can be used for adversarial detection for any OD algorithm.

\item \name supports the addition of new classes without any additional training, which is essential in retail where new items are added to the inventory on a daily basis.

\item The resources created in this research can be used by the research community to further investigate adversarial attacks in the retail domain, i.e., the Superstore dataset and the corresponding adversarial videos will be publicly available when the paper is published.

\end{itemize}

\section{Background}
Adversarial samples are real data samples that have been perturbed by an attacker to influence an ML model's prediction~\cite{chen2020survey,chakraborty2021survey}.
Numerous digital and physical adversarial attacks have been proposed~\cite{goodfellow2014explaining,madry2017towards,carlini2017towards,brown2017adversarial,hu2021naturalistic}, and recent studies have shown that such attacks, in the form of adversarial patches, can also target OD models~\cite{thys2019fooling,zolfi2021translucent,hu2021naturalistic,shapira2022denial}.
Since OD models are used for real-world tasks, those patches can even deceive the model in environments with high uncertainty~\cite{lee2019physical,zolfi2021translucent,hu2021naturalistic}.
An adversary that crafts an adversarial patch against an OD model may have one of three goals: 
\textit{i}) to prevent the OD model from detecting the presence of an object in a scene, i.e., perform a disappearance (hidden) attack~\cite{song2018physical,thys2019fooling,hu2021naturalistic,zolfi2021translucent};
\textit{ii}) to allow the OD model to successfully identify the object in a scene (correct bounding box) but cause it to be classified as a different object, i.e., perform a creation attack~\cite{zhu2021you};
or \textit{iii}) to cause the OD model to detect a non-existent object in a scene, i.e., perform an illusion attack~\cite{liu2018dpatch,lee2019physical}.
Examples of adversarial patch attacks that target OD models (which were used in \name's evaluation) include the DPatch~\cite{liu2018dpatch} and the illusion attack of Lee \& Kotler~\cite{lee2019physical}, which craft a targeted adversarial patch with minimal changes to the bounding box.

To successfully detect adversarial patches, \name utilizes four computer vision related techniques:
1) Object extraction~\cite{kirillov2020pointrend} -- the task of detecting and delineating the objects in a given scene, i.e., `cropping' the object presented in the scene and erasing its background;
2) Arbitrary style transfer~\cite{jing2019neural} -- an image manipulation technique that extracts
the style “characteristics” from a given style image and
blends them into a given input image;
3) Scale invariant feature transform (SIFT)~\cite{lowe2004distinctive} -- an explainable image matching technique that extracts a set of key points that represent the “essence” of an image, allowing the comparison of different images.
The key points are selected by examining their surrounding pixels' gradients after applying varied levels of blurring; and
4) The use of class prototypes~\cite{roscher2020explainable} -- an explainability technique that identifies data samples that best represent a class, i.e., the samples that are most related to a given class~\cite{molnar2020interpretable}.

\section{Related Work\label{Related_work}}
While adversarial detection in image classification has been extensively researched~\cite{xu2017feature,aldahdooh2022adversarial,chou2020sentinet,fidel2020explainability,yang2020ml}, only a few studies have been performed in the OD field. 
Moreover, it has been shown that adversarial patch attacks and detection techniques suited for image classification cannot be successfully transferred to the OD domain~\cite{liu2018dpatch,lu2017no,lu2017standard}.
Adversarial detection for OD can be divided into two categories: patch detection based on adversarial training (AT) and patch detection based on inconsistency comparison (IC).
The former is performed by adding adversarial samples to the OD model's training process so that the model becomes familiarized with the "adversarial" class, which will improve the detection rate~\cite{chiang2021adversarial,liu2022segment,xupatchzero}.
AT detectors can be applied both externally (the detector operates separately from the OD model) or internally (the detector is incorporated in the architecture of the OD model).
One example of an AT detector applied externally is Segment \& Complete (SAC)~\cite{liu2022segment}; SAC detects adversarial patches by training a separate segmentation model, which is used to detect and erase the adversarial patches from the scene.
In contrast, Ad-YOLO~\cite{ji2021adversarial} is an internal AT detector, which detects adversarial patches by adding them to the model's training set as an additional "adversarial" class.
The main limitation of AT detectors is that the detector's effectiveness is correlated with the attributes of the patches presented in the training set~\cite{zhang2019limitations}, i.e., the detector will have a lower detection rate for new patches.
In addition, none of the existing detectors provide a sufficient explanation for the alert raised.
% The main limitation of AT detectors is that the detector's effectiveness is higher on patches whose attributes are similar as those in the training set

The IC detection approach examines the similarity of the target model's prediction and the prediction of another ML model, which is referred to as the \textit{predictor}.
Any inconsistency between the two models' predictions will trigger an adversarial alert~\cite{xiang2021detectorguard}.
DetectorGuard~\cite{xiang2021detectorguard} is an example of a method employing IC; in this case, an instance segmentation model serves as the predictor.
DetectorGuard assumes that a patch cannot mislead an OD model and  instance segmentation model simultaneously. 
However, by relying on the output differences of those models, DetectorGuard is only effective against disappearance attacks and is not suitable for creation or illusion attacks that do not significantly alter the object's shape. 
Given this limitation, DetectorGuard is not a valid solution for the detection of such attacks, which are likely to be used in smart shopping cart systems.

\section{The Method}
\begin{figure*}[t]
\includegraphics[width=0.98\textwidth]{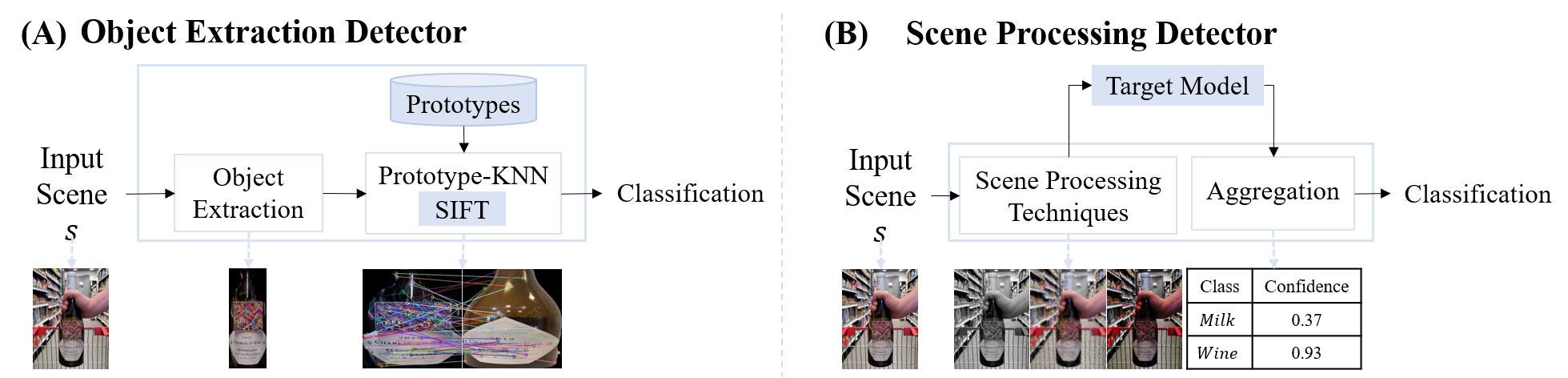}
\centering
\caption{X-Detect structure and components.}
\label{X-detect_structure}
\end{figure*}
\name's design is based on the assumption that the attacker crafts the adversarial patch for a specific attack environment (the target task is OD, and the input is preprocessed in a specific way), i.e., any change in the attack environment will harm the patch's capabilities.
\name starts by locating and classifying the main object in a given scene (which is the object most likely to be attacked) by using two explainable-by-design base detectors that change the attack environment. 
If there is a disagreement between the classification of \name and the target model, \name will raise an adversarial alert.
In this section, we introduce \name's components and structure (as illustrated in Figure~\ref{X-detect_structure}).
\name consists of two base detectors, the \textit{object extraction detector} (OED) and the \textit{scene processing detector} (SPD), each of which utilizes different scene manipulation techniques to neutralize the adversarial patch's effect on an object's classification.
These components can be used separately (by comparing the selected base detector's classification to the target model's classification) or as an ensemble to benefit from the advantages of both base detectors (by aggregating the outputs of both base detectors and comparing the result to the target model's classification).

The following notation is used:
Let $F$ be an OD model and $s$ be an input scene.
Let $F(s)=\{O_b,O_p,O_c\}$ be the output of $F$ for the main object in scene $s$, where $O_b$ is the object's bounding box, $O_c\in C$ is the classification of the object originating from the class set $C$, and $O_p\in[0,1]$ is the confidence of $F$ in the classification $O_c$.

\subsection{Object Extraction Detector \label{OED}}
The OED receives an input scene $s$ and outputs its classification for the main object in $s$.
First, the OED uses an object extraction model to eliminate the background noise from the main object in $s$.
As opposed to OD models, object extraction models use segmentation techniques that focus on the object's shape rather than on other properties.
Patch attacks on OD models change the object's classification without changing the object's outlines~\cite{liu2022segment}, therefore the patch will not affect the object extraction model's output.
Additionally, by using object extraction, the OED changes the assumed object surrounding by eliminating the scene's background, which may affect the final classification.
Then, the output of the object extraction model is classified by the \textit{prototype-KNN} classifier -- a customized KNN model.
KNN is an explainable-by-design algorithm, which, for a given sample, returns the $k$ closest samples according to a predefined proximity metric and uses majority voting to classify it.
Specifically, the \textit{prototype-KNN} chooses the $k$ closest neighbors from a predefined set of prototype samples $P$ from every class.
By changing the ML task to classification, the assumed attack environment has been changed.  
In addition, using prototypes as the neighbors guarantees that the set of neighbors will properly represent the different classes.
The \textit{prototype-KNN} proximity metric is based on the number of identical visual features shared by the two objects examined.
The visual features are extracted using SIFT~\cite{lowe2004distinctive}, and the object’s unique characteristics are represented by a set of key points.
The class of the prototype that has the highest number of matching key points with the examined object is selected.
The OED's functionality is presented in equation~\ref{equation:object extraction detector}:
\begin{equation}
\label{equation:object extraction detector}
OED(s) = P_{KNN}\bigg(\max_{p_i\in P}\Big|SIFT\big(OE(s),p_i\big)\Big|\bigg)
\end{equation}
where $P_{KNN}$ is the \textit{prototype-KNN}, $p_i \in P$ is a prototype sample, and $OE$ is the object extraction model.
The OED is considered explainable-by-design: 
\textit{i)} SIFT produces an explainable output that visually connects each matching point in the two scenes;
and \textit{ii)} the $K$ neighbors used for the classification can explain the decision of the \textit{prototype-KNN}.  

\subsection{Scene Processing Detector}
As the OED, the SPD receives an input scene $s$ and outputs its classification for the main object in $s$.
First, the SPD applies multiple image processing techniques on $s$. Then it feeds the processed scenes to the target model and aggregates the results into a classification.
The image processing techniques are applied to change the assumed OD pipeline by adding a preprocessing step, which would limit the patch's effect on the target model~\cite{thys2019fooling,hu2021naturalistic}.
The effect of the image processing techniques applied on the target model needs to be considered, i.e., a technique that harms the target model's performance on benign scenes (scenes without an adversarial patch) would be ineffective.
After the image processing techniques have been applied, the SPD feeds the processed scenes to the target model and receives the updated classifications.
The classifications of each processed scene are aggregated by selecting the class with the highest probability sum.
The SPD's functionality is presented in equation~\ref{equation:scene processing detector}:
\begin{equation}
\label{equation:scene processing detector}
SPD(s)  = Arg_{m \in SM} \Big(F\big(m(s)\big)\Big)
\end{equation}
where $m\in SM$ represents an image processing technique, and the main object's classification probability is aggregated from each processed image output by $Arg$.
The SPD is considered explainable-by-design, since it provides explanations for its alerts, i.e., every alert raised is accompanied by the processed scenes, which can be viewed as explanations-by-examples, i.e., samples that explain why X-Detect’s prediction changed.

\section{Evaluation}
\subsection {Datasets}
The following two datasets were used in the evaluation:
\textbf{Common Objects in Context (COCO) 2017}~\cite{lin2014microsoft} -- an OD benchmark containing 80 object classes and over 120K labeled images.\\
\noindent\textbf{Superstore} -- a dataset that we created, which is customized for the retail domain and the smart shopping cart use case.
The Superstore dataset contains 2,200 images (1,600 for training and 600 for testing), which are evenly distributed across 20 superstore products (classes).
Each image is annotated with a bounding box, the product's classification, and additional visual annotations (more information can be found in the supplementary material).
The Superstore dataset's main advantages are that all of the images were captured by cameras in a real smart cart setup (described in Section~\ref{Physical}) and it is highly diverse, i.e., the dataset can serve as a high-quality training set for related tasks.

\begin{figure}[t]
\centering
\begin{subfigure}[b]{0.44\textwidth}
    \includegraphics[width=\linewidth]{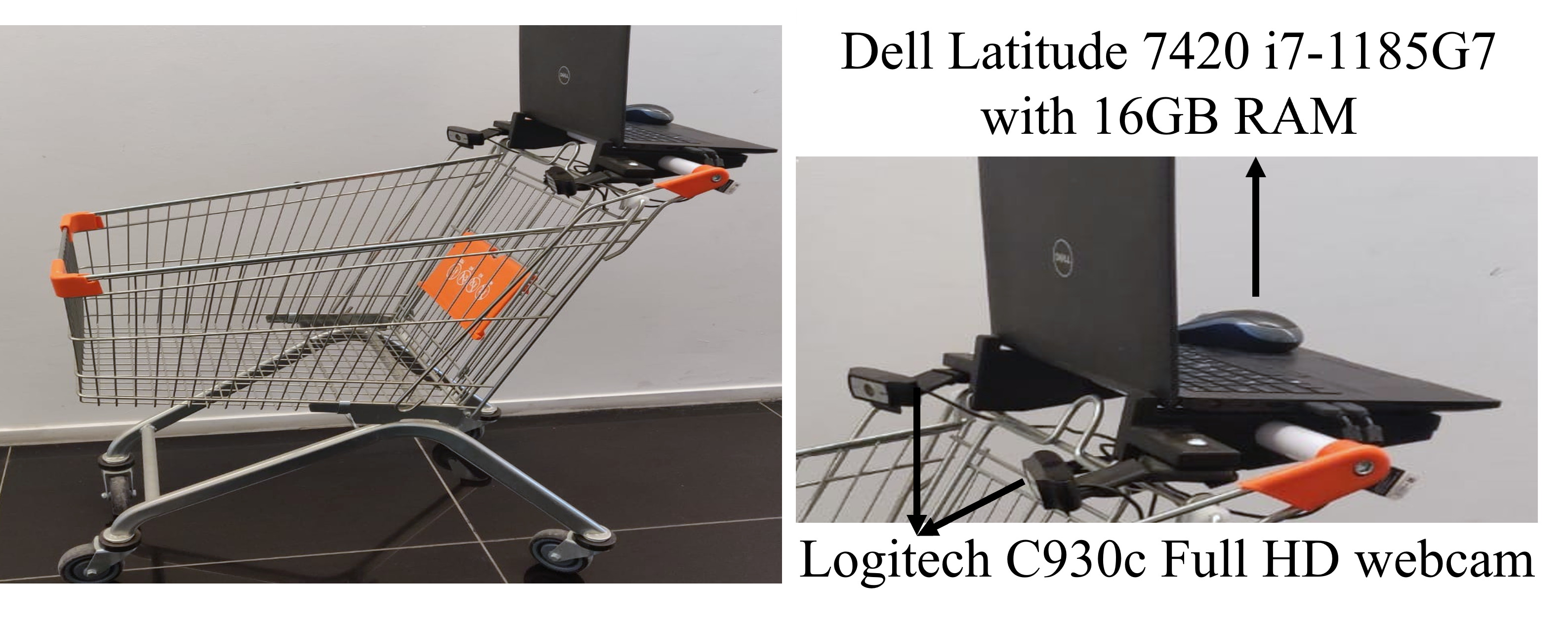}
    \centering
    \caption{Physical use case setup.}
    \label{Physical_setup}
\end{subfigure}
\begin{subfigure}[b]{0.55\textwidth}
    \scalebox{0.76}{
    \begin{tabular}{l|llll|}
    \cline{2-5}
     & \multicolumn{4}{c|}{Threat Model} \\ \cline{2-5} 
     & \multicolumn{1}{c|}{\begin{tabular}[c]{@{}c@{}}Defender \\ Settings \end{tabular}} & \multicolumn{1}{c|}{\begin{tabular}[c]{@{}c@{}}Model's \\ Weights \end{tabular}} & \multicolumn{1}{c|}{\begin{tabular}[c]{@{}c@{}}Model's \\ Architecture \end{tabular}} & \multicolumn{1}{c|}{\begin{tabular}[c]{@{}c@{}}ML \\ Algorithm\end{tabular}} \\ \hline
    \multicolumn{1}{|l|}{Adaptive Attack} & \multicolumn{1}{c|}{$ \checkmark $} & \multicolumn{1}{c|}{$ \checkmark $} & \multicolumn{1}{c|}{$ \checkmark $} & \multicolumn{1}{c|}{$ \checkmark $}  \\ \hline \hline
    \multicolumn{1}{|c|}{White-Box} & \multicolumn{1}{c|}{-} & \multicolumn{1}{c|}{$ \checkmark $} & \multicolumn{1}{c|}{$ \checkmark $} & \multicolumn{1}{c|}{$ \checkmark $} \\ \hline
    \multicolumn{1}{|c|}{Gray-Box} & \multicolumn{1}{c|}{-} & \multicolumn{1}{c|}{-} & \multicolumn{1}{c|}{$ \checkmark $} & \multicolumn{1}{c|}{$ \checkmark $}  \\ \hline
    \multicolumn{1}{|c|}{Model-Specific} & \multicolumn{1}{c|}{-} & \multicolumn{1}{c|}{-} & \multicolumn{1}{c|}{-} & \multicolumn{1}{c|}{$ \checkmark $} \\ \hline
    \multicolumn{1}{|c|}{Model-Agnostic} & \multicolumn{1}{c|}{-} & \multicolumn{1}{c|}{-} & \multicolumn{1}{c|}{-} & \multicolumn{1}{c|}{-} \\ \hline
    \end{tabular}}
    \caption{Attacker's knowledge about the attack scenarios examined.}
    \label{table:Threat model}
\end{subfigure}
\caption{Experimental evaluation settings: smart shopping cart setup (a), attack scenarios (b).}
\label{figtab}
\end{figure}

\subsection{\label{Physical}Evaluation Space}
\name was evaluated in two attack spaces: \textit{digital} and \textit{physical}.
In the \textit{digital space}, \name was evaluated under a digital attack with the COCO dataset.
In this use case, we used open-source pretrained OD models from the MMDetection framework's model zoo~\cite{mmdetection}.
We used 100 samples related to classes that are relevant to the smart shopping cart use case in the COCO dataset to craft two adversarial patches using the DPatch attack~\cite{liu2018dpatch}, each of which corresponds to a different target class - "Banana" or "Apple."
To create the adversarial samples, we placed the patches on 100 additional benign samples from four classes ("Banana," "Apple," "Orange," and "Pizza").
The test set  used to evaluate \name consisted of the 100 benign samples and their 100 adversarial samples.
We note that the adversarial patches were not placed on the corresponding benign class scenes, i.e., an "apple" patch was not placed on "apple" scenes and a "banana" patch was not placed on "banana" scenes.

In the \textit{physical space}, \name was evaluated in a real-world setup in which physical attacks were performed on real products from the Superstore dataset. 
For this evaluation, we designed a smart shopping cart using a shopping cart, two identical web cameras, and a personal computer (illustrated in Figure~\ref{Physical_setup}).
In this setup, the frames captured by the cameras were passed to a remote GPU server that stored the target OD model.
To craft the adversarial patches, we used samples of cheap products from the Superstore test set and divided them into two equally distributed data folds (15 samples from each class).
Each of the adversarial patches was crafted using one of the two folds.
The patches were crafted using the DPatch ~\cite{liu2018dpatch} and Lee \& Kotler ~\cite{lee2019physical} attacks (additional information is presented in supplementary material).
In total, we crafted 17 adversarial patches and recorded 1,700 \textrm{adversarial videos}, in which expensive products containing the adversarial patch were placed in the smart shopping cart.
The test set used to evaluate \name consisted of the \textrm{adversarial videos} along with an additional equal numbered of benign videos.
Those videos are available at.\footnote{drive.google.com/file/d/1VPbH0xVI7W-
ksPIIKdqxorNoTibfoeuT/view?usp=sharing}

\subsection{\label{attack_scenarios}Attack Scenarios}
Table~\ref{table:Threat model} presents the five attack scenarios evaluated and their corresponding threat models.
The attack scenarios can be categorized into two groups: non-adaptive attacks and adaptive attacks.
The former are attacks where the attacker threat model does not include knowledge about the defense approach, i.e., knowledge about the detection method used by the defender.
\name was evaluated on four non-adaptive attack scenarios that differ with regard to the attacker's knowledge: white-box (complete knowledge of the target model), gray-box (no knowledge of the target model's parameters), model-specific (knowledge on the ML algorithm used), and model agnostic (no knowledge on the target model). 
The patches used in these scenarios were crafted using the white-box scenario's target model; then they were used to simulate the attacker's knowledge in other scenarios (each scenario targets different models).  
We also carried out adaptive attacks~\cite{carlini2019evaluating}, i.e., the attacker threat model includes knowledge on the detection technique used and its parameters.
We designed three adaptive attacks that are based on the (LKpatch)~\cite{lee2019physical} attack of Lee \& Kotler, which is presented in equation~\ref{equation:DPatch attack}:
\begin{equation}
\label{equation:DPatch attack}
LK_{patch} = Clip\Big(P - \epsilon*sign\big(\bigtriangledown_p L(P,s,t,O_c,O_b)\big)\Big)
\end{equation}
where $P$ is the adversarial patch, $L$ is the target model's loss function, and $t$ are the transformations applied during training.
Each adaptive attack is designed according to \name's base detectors' settings: \textit{i)} using just the OED, \textit{ii)} using just the SPD, and \textit{iii)} using an ensemble of the two.
To adjust the LKpatch attack to consider the first setup (\textit{i}), we added a component to the attack's loss function that incorporates the core component of the OED - the \textit{prototype-KNN}.
The new loss component incorporates the SIFT algorithm's matching point outputs of the extracted object (with the patch) and the target class prototype.
The new loss component is presented in equation~\ref{equation:Adaptive object extraction attack}:
\begin{equation}
\label{equation:Adaptive object extraction attack}
\begin{split}
& OE_{SIFT} = -norm\Big(SIFT\big(OE(s,P),p_{target}\big)\Big)
\end{split}
\end{equation}
where $p_{target}$ is the target class prototype and $norm$ is a normalization function.

To adjust the LKpatch attack to incorporate the second setup (\textit{ii}), we added the image processing techniques used by the SPD to the transformations in the expectation over transformation functionality $t$.
The updated loss function is presented in equation~\ref{equation:Adaptive scene processing attack}:
\begin{equation}
\label{equation:Adaptive scene processing attack}
P_{ASP} = Clip\Big(P - \epsilon*sign\big(\bigtriangledown_p L(p,s,t,sp,O_c,O_b)\big)\Big)
\end{equation}
where $sp$ are the image processing techniques used.
To adjust the existing LKpatch attack to incorporate the last setup (\textit{iii}), we combined the two adaptive attacks described above, by adding $OE_{SIFT}$ to the $P_{ASP}$ loss.
Additional information regarding the adaptive attacks' intuition and implementation can be found in the supplementary material.

\subsection {Experimental Settings \label{Experimental_Settings}}
All of the experiments were performed on the CentOS Linux 7 (Core) operating system with an NVIDIA GeForce RTX 2080 Ti graphics card with 24GB of memory.
The code used in the experiments was written using Python 3.8.2, PyTorch 1.10.1, and NumPy 1.21.4 packages.
We used different target models, depending on the attack space and scenario in question.
In the digital space, only the white-box, model-specific, and model-agnostic attack scenarios were evaluated, since they are the most informative for this evaluation.
In the white-box scenario, the Faster R-CNN model with a ResNet-50-FPN backbone~\cite{ren2015faster,he2016deep} in PyTorch implementation was used.
In the model-specific scenario, three Faster R-CNN models were used - two with a ResNet-50-FPN backbone in Caffe implementation, in each one a different regression loss (IOU) was used, and the third used ResNet-101-FPN as a backbone.
In the model-agnostic scenario, the Cascade R-CNN~\cite{cai2019cascade} model and Grid R-CNN were used, each of which had a ResNet-50-FPN backbone. 
In the physical space evaluation, all five attack scenarios were evaluated.
In the adaptive and white-box scenarios, a Faster R-CNN model with a ResNet-50-FPN backbone was trained with the seed 42.
In the gray-box scenario, three Faster R-CNN models were trained (with the seeds 38, 40, 44) with a ResNet-50-FPN backbone.
In the model-agnostic scenario, a Cascade R-CNN model, Cascade RPN model~\cite{vu2019cascade}, and YOLOv3 model~\cite{redmon2018yolov3} were trained with the seed 42. The trained models are available at.\footnote{drive.google.com/file/d/1Dgk4AfqjJHWfTst-82XZeAZ-7XNypl8X/view?usp=sharing}

In the attacks performed, the learning rate was reduced automatically (on plateau), the batch size was set at one, and the patch was placed on the main object.
The patch size was set at 120*120 pixels in the digital space and 100*100 pixels in the physical space.
The transformations used in the LKpatch attack were brightness (between [0.8,1.6]) and random rotations.
Our code is also available at.\footnote{drive.google.com/file/d/1tOjcL8t3BlkrYNpgpgingwH6cCxL2Tc3 /view?usp=sharing}
All of the adversarial samples (i.e., scenes with an adversarial patch) were designed to meet the following three requirements
\textit{i}) the target OD model still identifies an object in the scene (there is a bounding box); 
\textit{ii}) the attack does not change the output's bounding box drastically;
and \textit{iii}) the attack changes the object classification to a predefined target class or group of classes (e.g., `cheap' products).
 
In the experiments, \name's components were set as follows.
The OED was initialized with 10 prototypes from each class.
To ensure that the prototypes were representative, we extracted the object associated with the prototype class.
The number of neighbors used by the \textit{prototype-KNN} was set to seven.
The scene processing detector used the following image processing techniques: blur (six in the digital space and 12 in the physical space), sharpen, random noise (0.35 in the physical space), darkness (0.1), and arbitrary style transfer.
The arbitrary style transfer technique used the evaluated scene served as both the input and the style image, i.e., the input scene was slightly changed.

The base detectors were evaluated separately and in the form of an ensemble.
Two types of ensembles were evaluated: 
\textit{i)} a majority voting (MV) ensemble, which sums the probabilities for each class and returns the class that received the highest sum;
and \textit{ii)} a 2-tier ensemble, which first applies the SPD, and if an alert is raised, the scene is passed to the OED.
To properly evaluate \name, we implemented two state-of-the-art adversarial detectors, Ad-YOLO~\cite{ji2021adversarial} and SAC ~\cite{liu2022segment} (Section~\ref{Related_work}), and compared their results to those of \name.
Both detectors were implemented according to their paper's details, and additional information is provided in the supplementary material.

\section{Experimental Results}\label{sec_results}
\begin{figure*}[t]
\includegraphics[width=0.99\textwidth]{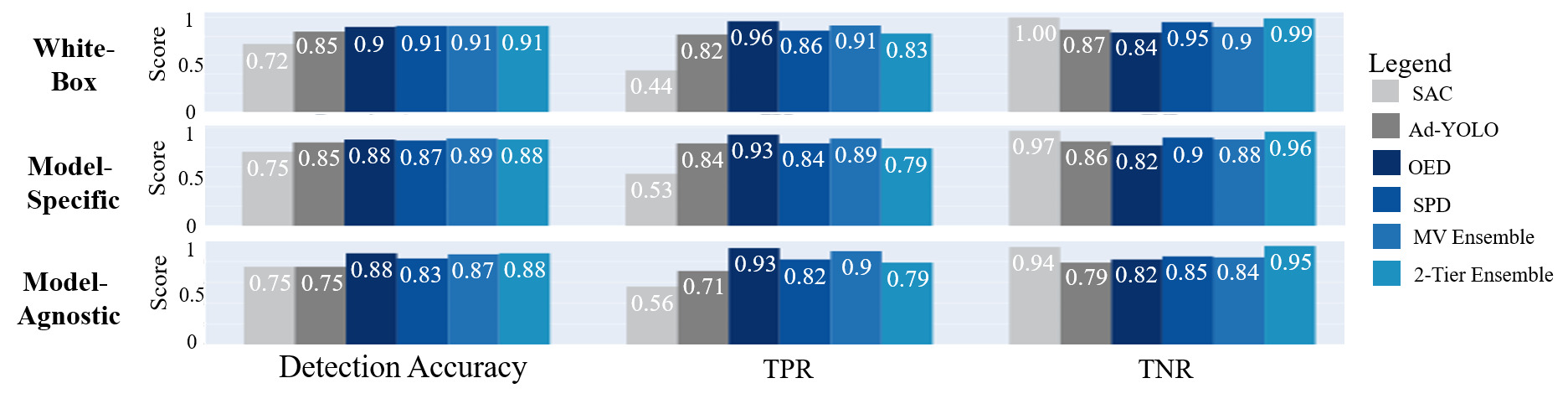}
\centering
\caption{Digital space evaluation results.}
\label{Digital use case evaluation results.}
\end{figure*}
The attacks used in the digital space (see Section~\ref{Physical}) reduced the mean average precision (mAP) in all attack scenarios by $~60\%$, i.e., the OD models' performance was substantially degraded.
In \name evaluation, only the successful adversarial samples were used.
There were $~95\%$, $~35\%$, and $~52\%$ successful adversarial samples respectively in the white-box, model-specific, and model-agnostic attack scenarios.
Further details are available in the supplementary material.
Figure~\ref{Digital use case evaluation results.} shows the \textit{digital space} evaluation results (performed on the COCO dataset) for \name, Ad-YOLO, and SAC detection for the white-box, model-specific, and model-agnostic attack scenarios.
The figure presents the detection accuracy (DA), true positive rate (TPR), and true negative rate (TNR) of \name components, Ad-YOLO, and SAC.
In the model-specific and model-agnostic attack scenarios, the results presented are the means obtained by each detector (see Section~\ref{attack_scenarios}) with a standard deviation of $0.034$ and $0.017$ respectively.
In the figure, we can see that \name outperformed Ad-YOLO and SAC on all of the performance metrics.
\name obtained the highest TPR with the OED; the highest TNR by the 2-tier ensemble (along with SAC); and the highest DA by the 2-tier ensemble.

The attacks used in the physical evaluation (described in Section~\ref{Physical}) decreased the models' performance substantially.
In \name evaluation, only the successful adversarial samples were used.
There were $80\%$, $80\%$, $74\%$, $33\%$, and $28\%$ successful adversarial samples in the adaptive, white-box, gray-box, model-specific, and model-agnostic scenarios respectively (additional information can be found in the supplementary material).
Table~\ref{Physical use case evaluation results.} presents the results of the \textit{physical space} evaluation (performed on the Superstore dataset) for \name, Ad-YOLO, and SAC for the white-box, gray-box, model-specific, and model-agnostic scenarios on six metrics: DA, TPR, TNR, false positive rate (FPR), false negative rate (FNR), and inference time (the detector's runtime for a single video).
The results of the gray-box, model-specific, and model-agnostic scenarios are the means obtained by each detector with a standard deviation of $0.032$, $0.044$, and $0.024$ respectively.
The results in the table show that \name in its different settings outperformed Ad-YOLO and SAC with the exception of the inference time metric ($\sim$0.4 seconds), which is discussed further in the supplementary material.
\name achieved the highest TPR by the OE detector; the highest TNR (along with SAC) by the 2-tier and MV ensembles; and the highest DA by the 2-tier ensemble.

\begin{table*}[t]
\caption{Physical space evaluation results.}
\label{Physical use case evaluation results.}
\begin{adjustbox}{width=0.80\textwidth,center}
\begin{tabular}{clcc|cccc|}
\cline{3-8}
\multicolumn{1}{l}{} & \multicolumn{1}{l|}{} & \multicolumn{2}{c|}{Existing Methods} & \multicolumn{4}{c|}{\name Evaluation} \\ \cline{2-8} 
\multicolumn{1}{l|}{} & \multicolumn{1}{l|}{Evaluation Metric} & \multicolumn{1}{l|}{Ad-YOLO} & \multicolumn{1}{l|}{SAC} & \multicolumn{1}{l|}{OED} & \multicolumn{1}{l|}{SPD} & \multicolumn{1}{l|}{MV Ensemble} & \multicolumn{1}{l|}{2-Tier Ensemble} \\ \hline \hline
\multicolumn{1}{|c|}{\multirow{4}{*}{White-Box}} & \multicolumn{1}{l|}{DA} & \multicolumn{1}{c|}{0.56} & \multicolumn{1}{c|}{0.54} & \multicolumn{1}{c|}{0.84} & \multicolumn{1}{c|}{0.91} & \multicolumn{1}{c|}{\textbf{0.96}} & 0.94 \\ \cline{2-8} 
\multicolumn{1}{|c|}{} & \multicolumn{1}{l|}{TPR} & \multicolumn{1}{c|}{0.13} & \multicolumn{1}{c|}{0.4} & \multicolumn{1}{c|}{\textbf{0.98}} & \multicolumn{1}{c|}{0.89} & \multicolumn{1}{c|}{0.92} & 0.88\\ \cline{2-8} 
\multicolumn{1}{|c|}{} & \multicolumn{1}{l|}{TNR} & \multicolumn{1}{c|}{\textbf{1.00}} & \multicolumn{1}{c|}{0.68} & \multicolumn{1}{c|}{0.70} & \multicolumn{1}{c|}{0.93} & \multicolumn{1}{c|}{\textbf{1.00}} & \textbf{1.00} \\ \cline{2-8}
\multicolumn{1}{|c|}{} & \multicolumn{1}{l|}{FPR} & \multicolumn{1}{c|}{\textbf{0.00}} & \multicolumn{1}{c|}{0.32} & \multicolumn{1}{c|}{0.3} & \multicolumn{1}{c|}{0.07} & \multicolumn{1}{c|}{\textbf{0.00}} & \textbf{0.00}\\ \cline{2-8}
\multicolumn{1}{|c|}{} & \multicolumn{1}{l|}{FNR} & \multicolumn{1}{c|}{0.87} & \multicolumn{1}{c|}{0.6} & \multicolumn{1}{c|}{\textbf{0.02}} & \multicolumn{1}{c|}{0.11} & \multicolumn{1}{c|}{0.08} & 0.12  \\ \hline \hline
% \multicolumn{1}{|c|}{} & \multicolumn{1}{l|}{Inference Time} & \multicolumn{1}{c|}{1.0} & \multicolumn{1}{c|}{\textbf{0.41}} & \multicolumn{1}{c|}{2.0} & \multicolumn{1}{c|}{2.6} & \multicolumn{1}{c|}{4.6} & 4.0 
\multicolumn{1}{|c|}{\multirow{4}{*}{Gray-Box}} & \multicolumn{1}{l|}{DA} & \multicolumn{1}{c|}{0.55} & \multicolumn{1}{c|}{0.51} & \multicolumn{1}{c|}{0.84} & \multicolumn{1}{c|}{0.85} & \multicolumn{1}{c|}{\textbf{0.92}} & 0.89 \\ \cline{2-8} 
\multicolumn{1}{|c|}{} & \multicolumn{1}{l|}{TPR} & \multicolumn{1}{c|}{0.11} & \multicolumn{1}{c|}{0.37} & \multicolumn{1}{c|}{\textbf{0.98}} & \multicolumn{1}{c|}{0.78} & \multicolumn{1}{c|}{0.88} & 0.78 \\ \cline{2-8} 
\multicolumn{1}{|c|}{} & \multicolumn{1}{l|}{TNR} & \multicolumn{1}{c|}{\textbf{1.00}} & \multicolumn{1}{c|}{0.62} & \multicolumn{1}{c|}{0.70} & \multicolumn{1}{c|}{0.93} & \multicolumn{1}{c|}{\textbf{1.00}} & \textbf{1.00} \\ \cline{2-8} 
\multicolumn{1}{|c|}{} & \multicolumn{1}{l|}{FPR} & \multicolumn{1}{c|}{\textbf{0.00}} & \multicolumn{1}{c|}{0.38} & \multicolumn{1}{c|}{0.3} & \multicolumn{1}{c|}{0.07} & \multicolumn{1}{c|}{\textbf{0.00}} & \textbf{0.00} \\ \cline{2-8}
\multicolumn{1}{|c|}{} & \multicolumn{1}{l|}{FNR} & \multicolumn{1}{c|}{0.89} & \multicolumn{1}{c|}{0.63} & \multicolumn{1}{c|}{\textbf{0.02}} & \multicolumn{1}{c|}{0.22} & \multicolumn{1}{c|}{0.12} & 0.22 \\ \hline \hline
% \multicolumn{1}{|c|}{} & \multicolumn{1}{l|}{Inference Time} & \multicolumn{1}{c|}{1.0} & \multicolumn{1}{c|}{\textbf{0.39}} & \multicolumn{1}{c|}{2.0} & \multicolumn{1}{c|}{2.6} & \multicolumn{1}{c|}{4.6} & 4.0 \\ \hline \hline
\multicolumn{1}{|c|}{\multirow{4}{*}{Model-Specific}} & \multicolumn{1}{l|}{DA} & \multicolumn{1}{c|}{0.56} & \multicolumn{1}{c|}{0.54} & \multicolumn{1}{c|}{0.82} & \multicolumn{1}{c|}{0.84} & \multicolumn{1}{c|}{\textbf{0.92}} & 0.88 \\ \cline{2-8} 
\multicolumn{1}{|c|}{} & \multicolumn{1}{l|}{TPR} & \multicolumn{1}{c|}{0.13} & \multicolumn{1}{c|}{0.39} & \multicolumn{1}{c|}{\textbf{0.94}} & \multicolumn{1}{c|}{0.74} & \multicolumn{1}{c|}{0.85} & 0.76 \\ \cline{2-8} 
\multicolumn{1}{|c|}{} & \multicolumn{1}{l|}{TNR} & \multicolumn{1}{c|}{\textbf{1.00}} & \multicolumn{1}{c|}{0.68} & \multicolumn{1}{c|}{0.70} & \multicolumn{1}{c|}{0.93} & \multicolumn{1}{c|}{\textbf{1.00}} & \textbf{1.00} \\ \cline{2-8}
\multicolumn{1}{|c|}{} & \multicolumn{1}{l|}{FPR} & \multicolumn{1}{c|}{\textbf{0.00}} & \multicolumn{1}{c|}{0.32} & \multicolumn{1}{c|}{0.3} & \multicolumn{1}{c|}{0.07} & \multicolumn{1}{c|}{\textbf{0.00}} & \textbf{0.00} \\ \cline{2-8} 
\multicolumn{1}{|c|}{} & \multicolumn{1}{l|}{FNR} & \multicolumn{1}{c|}{0.87} & \multicolumn{1}{c|}{0.61} & \multicolumn{1}{c|}{\textbf{0.06}} & \multicolumn{1}{c|}{0.26} & \multicolumn{1}{c|}{0.15} & 0.24 \\ \hline \hline
% \multicolumn{1}{|c|}{} & \multicolumn{1}{l|}{Inference Time} & \multicolumn{1}{c|}{1.0} & \multicolumn{1}{c|}{\textbf{0.45}} & \multicolumn{1}{c|}{2.0} & \multicolumn{1}{c|}{2.6} & \multicolumn{1}{c|}{4.6} & 4.0 \\ \hline \hline
\multicolumn{1}{|c|}{\multirow{4}{*}{Model-Agnostic}} & \multicolumn{1}{l|}{DA} & \multicolumn{1}{c|}{0.45} & \multicolumn{1}{c|}{0.52} & \multicolumn{1}{c|}{0.82} & \multicolumn{1}{c|}{0.85} & \multicolumn{1}{c|}{\textbf{0.90}} & 0.88 \\ \cline{2-8} 
\multicolumn{1}{|c|}{} & \multicolumn{1}{l|}{TPR} & \multicolumn{1}{c|}{0.10} & \multicolumn{1}{c|}{0.45} & \multicolumn{1}{c|}{\textbf{0.95}} & \multicolumn{1}{c|}{0.77} & \multicolumn{1}{c|}{0.85} & 0.75 \\ \cline{2-8} 
\multicolumn{1}{|c|}{} & \multicolumn{1}{l|}{TNR} & \multicolumn{1}{c|}{0.90} & \multicolumn{1}{c|}{0.58} & \multicolumn{1}{c|}{0.69} & \multicolumn{1}{c|}{0.93} & \multicolumn{1}{c|}{0.96} & \textbf{0.98} \\ \cline{2-8}
\multicolumn{1}{|c|}{} & \multicolumn{1}{l|}{FPR} & \multicolumn{1}{c|}{0.10} & \multicolumn{1}{c|}{0.42} & \multicolumn{1}{c|}{0.31} & \multicolumn{1}{c|}{0.07} & \multicolumn{1}{c|}{0.04} & \textbf{0.02} \\ \cline{2-8} 
\multicolumn{1}{|c|}{} & \multicolumn{1}{l|}{FNR} & \multicolumn{1}{c|}{0.90} & \multicolumn{1}{c|}{0.55} & \multicolumn{1}{c|}{\textbf{0.05}} & \multicolumn{1}{c|}{0.23} & \multicolumn{1}{c|}{0.15} & 0.25 \\ \hline \hline
% \multicolumn{1}{|c|}{} & \multicolumn{1}{l|}{Inference Time} & \multicolumn{1}{c|}{1.0} & \multicolumn{1}{c|}{\textbf{0.4}} & \multicolumn{1}{c|}{2.0} & \multicolumn{1}{c|}{2.6} & \multicolumn{1}{c|}{4.6} & 4.0 \\ \hline \hline
\end{tabular}
\end{adjustbox}
\end{table*}

\begin{table*}[t]
\caption{Adaptive attack detection. 
% Gray cells indicate an adaptive attack type targeting its corresponding detector.}
Gray cells are the targeted detector's results.}
\label{Adaptive attack detection}
\begin{adjustbox}{width=0.80\textwidth,center}
\begin{tabular}{ccl|cccc|}
\cline{4-7}
\multicolumn{1}{l}{} & \multicolumn{1}{l}{} &  & \multicolumn{4}{c|}{X-Detect Evaluation} \\ \hline
\multicolumn{1}{|c|}{\begin{tabular}[c]{@{}c@{}}Adaptive Attack Type\end{tabular}} & \multicolumn{1}{c|}{\begin{tabular}[c]{@{}c@{}}Attack Success\end{tabular}} & \multicolumn{1}{c|}{\begin{tabular}[c]{@{}c@{}}Evaluation Metric\end{tabular}} & \multicolumn{1}{c|}{\begin{tabular}[c]{@{}c@{}}OED\end{tabular}} & \multicolumn{1}{c|}{\begin{tabular}[c]{@{}c@{}}SPD\end{tabular}} & \multicolumn{1}{c|}{MV Ensemble} & \begin{tabular}[c]{@{}c@{}}2-Tier Ensemble\end{tabular} \\ \hline \hline
\multicolumn{1}{|c|}{} & \multicolumn{1}{c|}{} & TPR & \multicolumn{1}{c|}{\cellcolor{Gray}\textbf{1.00}} & \multicolumn{1}{c|}{0.96} & \multicolumn{1}{c|}{0.92} & 0.96 \\ \cline{3-7} 
\multicolumn{1}{|c|}{\multirow{-2}{*}{Object Extraction}} & \multicolumn{1}{c|}{\multirow{-2}{*}{99\%}} & TNR & \multicolumn{1}{c|}{\cellcolor{Gray}\textbf{0.70}} & \multicolumn{1}{c|}{0.93} & \multicolumn{1}{c|}{1.00} & 1.00 \\ \hline \hline
\multicolumn{1}{|c|}{} & \multicolumn{1}{c|}{} & TPR & \multicolumn{1}{c|}{\textbf{1.00}} & \multicolumn{1}{c|}{\cellcolor{Gray}0.90} & \multicolumn{1}{c|}{0.96} & 0.93 \\ \cline{3-7}  
\multicolumn{1}{|c|}{\multirow{-2}{*}{Scene Processing}} & \multicolumn{1}{c|}{\multirow{-2}{*}{89\%}} & TNR & \multicolumn{1}{c|}{\textbf{0.70}} & \multicolumn{1}{c|}{\cellcolor{Gray}0.93} & \multicolumn{1}{c|}{1.00} & 1.00 \\ \hline \hline
\multicolumn{1}{|c|}{} & \multicolumn{1}{c|}{} & TPR & \multicolumn{1}{c|}{\textbf{1.00}} & \multicolumn{1}{c|}{0.90} & \multicolumn{1}{c|}{\cellcolor{Gray}0.95} & \cellcolor{Gray}0.90 \\ \cline{3-7} 
\multicolumn{1}{|c|}{\multirow{-2}{*}{Ensemble}} & \multicolumn{1}{c|}{\multirow{-2}{*}{53\%}} & TNR & \multicolumn{1}{c|}{\textbf{0.70}} & \multicolumn{1}{c|}{0.93} & \multicolumn{1}{c|}{\cellcolor{Gray}1.00} & \cellcolor{Gray}1.00 \\ \hline \hline
\end{tabular}
\end{adjustbox}
\end{table*}

% \name was evaluated also on the adaptive attack scenario (Section~\ref{attack_scenarios}) where the attacker has complete knowledge of \name settings.
We also evaluated \name's performance in the adaptive attack scenario.
Adaptive attacks require the addition of a component that enables the attack to evade the known defense mechanism, making them more challenging to perform (see Section \ref{attack_scenarios}).
When using \name's parameters in the adaptive attacks' optimization process, the attacks did not converge, i.e., the attacks failed to produce a patch that deceived the target model while evading the defense method.
Therefore, to evaluate \name when faced with a proper attack, the parameters used were relaxed (see Section \ref{Experimental_Settings} and the supplementary material).
Table~\ref{Adaptive attack detection} presents \name's TPR and TNR in the adaptive attack scenario.
The results indicate that while these adaptive patches partially succeeded in deceiving the target model, they did not succeed in evading \name, i.e., \name successfully detected most of the adversarial samples while maintaining a TPR of at least $92\%$.
This shows that evading \name is not possible, even when relaxing some of its parameters, and without doing so, the adaptive attacks did not converge.
Therefore, \name is successful in detecting adversarial patches in the adaptive attack scenario (more information can be found in the supplementary material).
Figure \ref{X-Detect explanation} presents examples of \name’s explainable outputs, which can be used to justify the alert raised. 

\section{Discussion}
During the evaluation, several interesting attack behaviors were observed.
When analyzing the attack's behavior in the crafting phase, we observed that the location of the adversarial patch influenced the attack's success, i.e., when the adversarial patch was placed on the attacked object, the attack's success rate improved.
Furthermore, in analyzing the attacks in the physical space, we observed their sensitivity to the attacker's behavior, such as the angle, speed of item insertion, and patch size.

When analyzing the attacks in the different attack scenarios, we observed that as the knowledge available for the attacker decreased, the number of successful adversarial samples also decreased.
The reason for this is that the shift between attack scenarios relies on the patch's transferability.
Therefore, the adversarial samples that succeed in the most restricted attack scenario (model-agnostic) can be considered the most robust.
Those samples would be harder to detect, as reflected in the performance of \name in its different settings. 
However, all the results show the detection of a single attack; in the shopping cart use case, an attacker would likely try to steal as many items as possible.
% When considering the occurrence of a repeated attack and its sensitivity to the attacker's behavior, \name's ability to expose the attacker would increase exponentially.  
In the case of a repeated attack, \name's ability to expose the attacker would improve exponentially.

In the evaluation, four detection approaches were used: only the OED, only the SPD, an MV ensemble, and a 2-tier ensemble, however the question of which approach is the most suitable for the adversarial detection task remains unanswered.
Each of the approaches showed different strengths and weaknesses:
\textit{i)} the OED approach managed to detect most of the adversarial samples (high TPR) yet raised alerts on benign samples (high FPR);
\textit{ii)} the SPD approach detected fewer adversarial samples than the OED (lower TPR) yet raised fewer alerts on benign samples (lower FPR);
\textit{iii)} the MV ensemble reduced the gap between the two base-detectors' performance (higher TPR and lower FPR) yet had a longer inference time;
and \textit{iv)} the 2-tier ensemble reduced the MV ensemble's inference time and improved the identification of benign samples (higher TNR and lower FPR) yet detected fewer adversarial samples (lower TPR).
Therefore, the selection of the best approach depends on the use case.
In the retail domain, it can be assumed that:
\textit{i)} most customers would not use an adversarial patch to shoplift;
\textit{ii)} wrongly accusing a customer of shoplifting would result in a dissatisfied customer, unlikely to return to the store, and harm the company's reputation;
\textit{iii)} short inference time is vital in real-time applications like the smart shopping cart.
Therefore, a company that places more value on the customer experience would prefer the 2-tier ensemble approach, while a company that prioritizes revenue above all would prefer the OED or MV ensemble approach.

\begin{figure*}[t]
\includegraphics[width=1.00\textwidth]{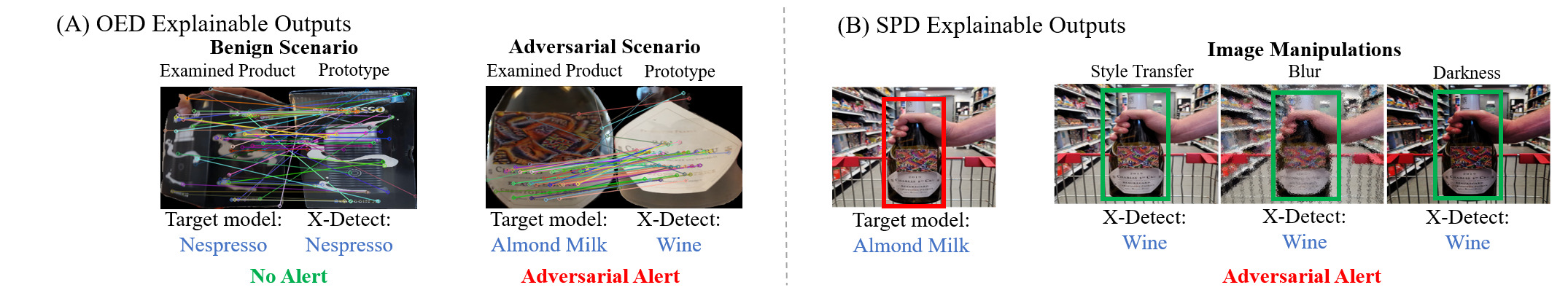}
\centering
\caption{X-Detect's explainable output. 
(A) The OED output - matching points between the examined product and its most similar prototype in the benign and adversarial scenarios. 
(B) The SPD output - the classification result for the given scene and various image manipulations.
}
\label{X-Detect explanation}
\end{figure*}

\section{Conclusion and Future Work}
In this paper, we presented \name, a novel adversarial detector for OD models which is suitable for real-life settings.
In contrast to existing methods, \name is capable of:
\textit{i)} identifying adversarial samples in near real-time;
\textit{ii)} providing explanations for the alerts raised; and \textit{iii)} handling new attacks.
Our evaluation in the digital and physical spaces, which was performed using a smart shopping cart setup, demonstrated that \name outperforms existing methods in the task of distinguishing between benign and adversarial scenes in the four attack scenarios examined while maintaining a 0\% FPR.
Furthermore, we demonstrated \name effectively under adaptive attacks.
Future work may include applying \name in different domains (security, autonomous vehicles, etc.) and expanding it for the detection of other kinds of attacks, such as physical backdoor attacks against ODs.

% \bibliographystyle{acm}
% \bibliography{Arxiv2023/Paper}

% \section*{References}
% \bibliographystyle{acm}
% \bibliography{Arxiv2023/Paper.bib}
% % \bibliography{Arxiv2023/Paper.bbl}

\begin{thebibliography}{10}

\bibitem{aldahdooh2022adversarial}
{\sc Aldahdooh, A., Hamidouche, W., Fezza, S.~A., and D{\'e}forges, O.}
\newblock Adversarial example detection for dnn models: A review and
  experimental comparison.
\newblock {\em Artificial Intelligence Review\/} (2022), 1--60.

\bibitem{ASPD2022}
{\sc Amazon}.
\newblock Amazon shoplifting punishment detection 2022.
\newblock
  \url{theverge.com/2018/1/22/16920784/amazon-go-cashier-less-grocery-store-seattle-shoplifting-punishment-detection}.

\bibitem{brown2017adversarial}
{\sc Brown, T.~B., Man{\'e}, D., Roy, A., Abadi, M., and Gilmer, J.}
\newblock Adversarial patch.
\newblock {\em arXiv preprint arXiv:1712.09665\/} (2017).

\bibitem{cai2021rethinking}
{\sc Cai, Y., Wen, L., Zhang, L., Du, D., and Wang, W.}
\newblock Rethinking object detection in retail stores.
\newblock In {\em Proceedings of the AAAI Conference on Artificial
  Intelligence\/} (2021), vol.~35, pp.~947--954.

\bibitem{cai2019cascade}
{\sc Cai, Z., and Vasconcelos, N.}
\newblock Cascade r-cnn: high quality object detection and instance
  segmentation.
\newblock {\em IEEE transactions on pattern analysis and machine intelligence
  43}, 5 (2019), 1483--1498.

\bibitem{carlini2019evaluating}
{\sc Carlini, N., Athalye, A., Papernot, N., Brendel, W., Rauber, J., Tsipras,
  D., Goodfellow, I., Madry, A., and Kurakin, A.}
\newblock On evaluating adversarial robustness.
\newblock {\em arXiv preprint arXiv:1902.06705\/} (2019).

\bibitem{carlini2017towards}
{\sc Carlini, N., and Wagner, D.}
\newblock Towards evaluating the robustness of neural networks.
\newblock In {\em 2017 ieee symposium on security and privacy (sp)\/} (2017),
  IEEE, pp.~39--57.

\bibitem{chakraborty2021survey}
{\sc Chakraborty, A., Alam, M., Dey, V., Chattopadhyay, A., and Mukhopadhyay,
  D.}
\newblock A survey on adversarial attacks and defences.
\newblock {\em CAAI Transactions on Intelligence Technology 6}, 1 (2021),
  25--45.

\bibitem{mmdetection}
{\sc Chen, K., Wang, J., Pang, J., Cao, Y., Xiong, Y., Li, X., Sun, S., Feng,
  W., Liu, Z., Xu, J., Zhang, Z., Cheng, D., Zhu, C., Cheng, T., Zhao, Q., Li,
  B., Lu, X., Zhu, R., Wu, Y., Dai, J., Wang, J., Shi, J., Ouyang, W., Loy,
  C.~C., and Lin, D.}
\newblock {MMDetection}: Open mmlab detection toolbox and benchmark.
\newblock {\em arXiv preprint arXiv:1906.07155\/} (2019).

\bibitem{chen2020survey}
{\sc Chen, K., Zhu, H., Yan, L., and Wang, J.}
\newblock A survey on adversarial examples in deep learning.
\newblock {\em Journal on Big Data 2}, 2 (2020), 71.

\bibitem{cheng2021fashion}
{\sc Cheng, W.-H., Song, S., Chen, C.-Y., Hidayati, S.~C., and Liu, J.}
\newblock Fashion meets computer vision: A survey.
\newblock {\em ACM Computing Surveys (CSUR) 54}, 4 (2021), 1--41.

\bibitem{chiang2021adversarial}
{\sc Chiang, P.-H., Chan, C.-S., and Wu, S.-H.}
\newblock Adversarial pixel masking: A defense against physical attacks for
  pre-trained object detectors.
\newblock In {\em Proceedings of the 29th ACM International Conference on
  Multimedia\/} (2021), pp.~1856--1865.

\bibitem{chou2020sentinet}
{\sc Chou, E., Tramer, F., and Pellegrino, G.}
\newblock Sentinet: Detecting localized universal attacks against deep learning
  systems.
\newblock In {\em 2020 IEEE Security and Privacy Workshops (SPW)\/} (2020),
  IEEE, pp.~48--54.

\bibitem{nrss2022}
{\sc Federation, N.~R.}
\newblock National retail security survey 2022.
\newblock \url{https://nrf.com/research/national-retail-security-survey-2022}.

\bibitem{fidel2020explainability}
{\sc Fidel, G., Bitton, R., and Shabtai, A.}
\newblock When explainability meets adversarial learning: Detecting adversarial
  examples using shap signatures.
\newblock In {\em 2020 international joint conference on neural networks
  (IJCNN)\/} (2020), IEEE, pp.~1--8.

\bibitem{FSR2022}
{\sc Forbes}.
\newblock Forbes shoplifting report 2022.
\newblock
  \url{https://forbes.com/sites/jiawertz/2022/11/20/shoplifting-has-become-a-100-billion-problem-for-retailers/?sh=679b9a282d62}.

\bibitem{fuchs2019towards}
{\sc Fuchs, K., Grundmann, T., and Fleisch, E.}
\newblock Towards identification of packaged products via computer vision:
  Convolutional neural networks for object detection and image classification
  in retail environments.
\newblock In {\em Proceedings of the 9th International Conference on the
  Internet of Things\/} (2019), pp.~1--8.

\bibitem{goodfellow2014explaining}
{\sc Goodfellow, I.~J., Shlens, J., and Szegedy, C.}
\newblock Explaining and harnessing adversarial examples.
\newblock {\em arXiv preprint arXiv:1412.6572\/} (2014).

\bibitem{green2021super}
{\sc Green, K.~M.}
\newblock {\em Super-Big Market-Data: A Case Study, Walkthrough Approach to
  Amazon Go Cashierless Convenience Stores}.
\newblock PhD thesis, University of Illinois at Chicago, 2021.

\bibitem{he2016deep}
{\sc He, K., Zhang, X., Ren, S., and Sun, J.}
\newblock Deep residual learning for image recognition.
\newblock In {\em Proceedings of the IEEE conference on computer vision and
  pattern recognition\/} (2016), pp.~770--778.

\bibitem{hu2021naturalistic}
{\sc Hu, Y.-C.-T., Kung, B.-H., Tan, D.~S., Chen, J.-C., Hua, K.-L., and Cheng,
  W.-H.}
\newblock Naturalistic physical adversarial patch for object detectors.
\newblock In {\em Proceedings of the IEEE/CVF International Conference on
  Computer Vision\/} (2021), pp.~7848--7857.

\bibitem{ji2021adversarial}
{\sc Ji, N., Feng, Y., Xie, H., Xiang, X., and Liu, N.}
\newblock Adversarial yolo: Defense human detection patch attacks via detecting
  adversarial patches.
\newblock {\em arXiv preprint arXiv:2103.08860\/} (2021).

\bibitem{jing2019neural}
{\sc Jing, Y., Yang, Y., Feng, Z., Ye, J., Yu, Y., and Song, M.}
\newblock Neural style transfer: A review.
\newblock {\em IEEE transactions on visualization and computer graphics 26}, 11
  (2019), 3365--3385.

\bibitem{kalli2021effective}
{\sc Kalli, S., Suresh, T., Prasanth, A., Muthumanickam, T., and Mohanram, K.}
\newblock An effective motion object detection using adaptive background
  modeling mechanism in video surveillance system.
\newblock {\em Journal of Intelligent \& Fuzzy Systems}, Preprint (2021),
  1--13.

\bibitem{khatab2021vulnerable}
{\sc Khatab, E., Onsy, A., Varley, M., and Abouelfarag, A.}
\newblock Vulnerable objects detection for autonomous driving: A review.
\newblock {\em Integration 78\/} (2021), 36--48.

\bibitem{kirillov2020pointrend}
{\sc Kirillov, A., Wu, Y., He, K., and Girshick, R.}
\newblock Pointrend: Image segmentation as rendering.
\newblock In {\em Proceedings of the IEEE/CVF conference on computer vision and
  pattern recognition\/} (2020), pp.~9799--9808.

\bibitem{lee2019physical}
{\sc Lee, M., and Kolter, Z.}
\newblock On physical adversarial patches for object detection.
\newblock {\em arXiv preprint arXiv:1906.11897\/} (2019).

\bibitem{lin2014microsoft}
{\sc Lin, T.-Y., Maire, M., Belongie, S., Hays, J., Perona, P., Ramanan, D.,
  Doll{\'a}r, P., and Zitnick, C.~L.}
\newblock Microsoft coco: Common objects in context.
\newblock In {\em European conference on computer vision\/} (2014), Springer,
  pp.~740--755.

\bibitem{liu2022segment}
{\sc Liu, J., Levine, A., Lau, C.~P., Chellappa, R., and Feizi, S.}
\newblock Segment and complete: Defending object detectors against adversarial
  patch attacks with robust patch detection.
\newblock In {\em Proceedings of the IEEE/CVF Conference on Computer Vision and
  Pattern Recognition\/} (2022), pp.~14973--14982.

\bibitem{liu2018dpatch}
{\sc Liu, X., Yang, H., Liu, Z., Song, L., Li, H., and Chen, Y.}
\newblock Dpatch: An adversarial patch attack on object detectors.
\newblock {\em arXiv preprint arXiv:1806.02299\/} (2018).

\bibitem{lowe2004distinctive}
{\sc Lowe, D.~G.}
\newblock Distinctive image features from scale-invariant keypoints.
\newblock {\em International journal of computer vision 60}, 2 (2004), 91--110.

\bibitem{lu2017no}
{\sc Lu, J., Sibai, H., Fabry, E., and Forsyth, D.}
\newblock No need to worry about adversarial examples in object detection in
  autonomous vehicles.
\newblock {\em arXiv preprint arXiv:1707.03501\/} (2017).

\bibitem{lu2017standard}
{\sc Lu, J., Sibai, H., Fabry, E., and Forsyth, D.}
\newblock Standard detectors aren't (currently) fooled by physical adversarial
  stop signs.
\newblock {\em arXiv preprint arXiv:1710.03337\/} (2017).

\bibitem{madry2017towards}
{\sc Madry, A., Makelov, A., Schmidt, L., Tsipras, D., and Vladu, A.}
\newblock Towards deep learning models resistant to adversarial attacks.
\newblock {\em arXiv preprint arXiv:1706.06083\/} (2017).

\bibitem{melek2017survey}
{\sc Melek, C.~G., Sonmez, E.~B., and Albayrak, S.}
\newblock A survey of product recognition in shelf images.
\newblock In {\em 2017 International Conference on Computer Science and
  Engineering (UBMK)\/} (2017), IEEE, pp.~145--150.

\bibitem{molnar2020interpretable}
{\sc Molnar, C.}
\newblock {\em Interpretable machine learning}.
\newblock Lulu. com, 2020.

\bibitem{oh2020implementation}
{\sc Oh, J.-S., and Chun, I.-G.}
\newblock Implementation of smart shopping cart using object detection method
  based on deep learning.
\newblock {\em Journal of the Korea Academia-Industrial cooperation Society
  21}, 7 (2020), 262--269.

\bibitem{redmon2018yolov3}
{\sc Redmon, J., and Farhadi, A.}
\newblock Yolov3: An incremental improvement.
\newblock {\em arXiv preprint arXiv:1804.02767\/} (2018).

\bibitem{ren2015faster}
{\sc Ren, S., He, K., Girshick, R., and Sun, J.}
\newblock Faster r-cnn: Towards real-time object detection with region proposal
  networks.
\newblock {\em Advances in neural information processing systems 28\/} (2015).

\bibitem{roscher2020explainable}
{\sc Roscher, R., Bohn, B., Duarte, M.~F., and Garcke, J.}
\newblock Explainable machine learning for scientific insights and discoveries.
\newblock {\em Ieee Access 8\/} (2020), 42200--42216.

\bibitem{santra2019comprehensive}
{\sc Santra, B., and Mukherjee, D.~P.}
\newblock A comprehensive survey on computer vision based approaches for
  automatic identification of products in retail store.
\newblock {\em Image and Vision Computing 86\/} (2019), 45--63.

\bibitem{shapira2022denial}
{\sc Shapira, A., Zolfi, A., Demetrio, L., Biggio, B., and Shabtai, A.}
\newblock Denial-of-service attack on object detection model using universal
  adversarial perturbation.
\newblock {\em arXiv preprint arXiv:2205.13618\/} (2022).

\bibitem{song2018physical}
{\sc Song, D., Eykholt, K., Evtimov, I., Fernandes, E., Li, B., Rahmati, A.,
  Tramer, F., Prakash, A., and Kohno, T.}
\newblock Physical adversarial examples for object detectors.
\newblock In {\em 12th USENIX workshop on offensive technologies (WOOT 18)\/}
  (2018).

\bibitem{song2021object}
{\sc Song, Q., Li, S., Bai, Q., Yang, J., Zhang, X., Li, Z., and Duan, Z.}
\newblock Object detection method for grasping robot based on improved yolov5.
\newblock {\em Micromachines 12}, 11 (2021), 1273.

\bibitem{thys2019fooling}
{\sc Thys, S., Van~Ranst, W., and Goedem{\'e}, T.}
\newblock Fooling automated surveillance cameras: adversarial patches to attack
  person detection.
\newblock In {\em Proceedings of the IEEE/CVF conference on computer vision and
  pattern recognition workshops\/} (2019), pp.~0--0.

\bibitem{vu2019cascade}
{\sc Vu, T., Jang, H., Pham, T.~X., and Yoo, C.}
\newblock Cascade rpn: Delving into high-quality region proposal network with
  adaptive convolution.
\newblock {\em Advances in neural information processing systems 32\/} (2019).

\bibitem{xiang2021detectorguard}
{\sc Xiang, C., and Mittal, P.}
\newblock Detectorguard: Provably securing object detectors against localized
  patch hiding attacks.
\newblock In {\em Proceedings of the 2021 ACM SIGSAC Conference on Computer and
  Communications Security\/} (2021), pp.~3177--3196.

\bibitem{xupatchzero}
{\sc Xu, K., Xiao, Y., Zheng, Z., Cai, K., and Nevatia, R.}
\newblock Patchzero: Defending against adversarial patch attacks by detecting
  and zeroing the patch.

\bibitem{xu2017feature}
{\sc Xu, W., Evans, D., and Qi, Y.}
\newblock Feature squeezing: Detecting adversarial examples in deep neural
  networks.
\newblock {\em arXiv preprint arXiv:1704.01155\/} (2017).

\bibitem{yang2020ml}
{\sc Yang, P., Chen, J., Hsieh, C.-J., Wang, J.-L., and Jordan, M.}
\newblock Ml-loo: Detecting adversarial examples with feature attribution.
\newblock In {\em Proceedings of the AAAI Conference on Artificial
  Intelligence\/} (2020), vol.~34, pp.~6639--6647.

\bibitem{zhang2019limitations}
{\sc Zhang, H., Chen, H., Song, Z., Boning, D., Dhillon, I.~S., and Hsieh,
  C.-J.}
\newblock The limitations of adversarial training and the blind-spot attack.
\newblock {\em arXiv preprint arXiv:1901.04684\/} (2019).

\bibitem{zhu2021you}
{\sc Zhu, Z., Su, H., Liu, C., Xiang, W., and Zheng, S.}
\newblock You cannot easily catch me: A low-detectable adversarial patch for
  object detectors.
\newblock {\em arXiv preprint arXiv:2109.15177\/} (2021).

\bibitem{zolfi2021translucent}
{\sc Zolfi, A., Kravchik, M., Elovici, Y., and Shabtai, A.}
\newblock The translucent patch: A physical and universal attack on object
  detectors.
\newblock In {\em Proceedings of the IEEE/CVF Conference on Computer Vision and
  Pattern Recognition\/} (2021), pp.~15232--15241.

\end{thebibliography}
% \input{Arxiv2023/Paper.bbl}

%%%%%%%%%%%%%%%%%%%%%%%%%%%%%%%%%%%%%%%%%%%%%%%%%%%%%%%%%%%%

\end{document}

\maketitle
\section{Abstract}
In this supplementary material, we provide additional information related to the evaluation experiments and results.
First, we elaborate on the additional information gathered during the evaluation settings, this includes a comprehensive description of the smart shopping cart setup and its characteristics. 
Next, we present additional information regarding the properties and statistics of the Superstore dataset used in our evaluation and provide supplementary details concerning the crafted patches used in the experiments. 
Lastly, we introduce additional details regarding the adaptive attack implementation and results and elaborate on our experimental results.

\section{Evaluation Settings Additional Information}

\subsection{Smart Shopping Cart Setup}

To perform the evaluation in the physical space, we designed a smart shopping cart setup using a regular shopping cart, two identical web cameras, and a personal computer.
The cart's volume was 100 liters, and the cameras used were the Logitech C930c Full HD Webcam with a resolution of 1080p. 
In addition, we designed and 3D printed a plugin on which to place the computer and cameras on the cart. 
The stand and smart shopping cart (and its dimensions) are presented in Figure~\ref{Physical_setup}.
To perform real-time detection of products as they are being inserted into the cart, a client-server architecture was employed.
The model (the server) detects products and sends the results to the customer display (the client).
The frames are captured by the cameras on the client side and then passed to a remote GPU server that stores the target object detection model.
The server processes the frames and feeds them to the object detection model.
Once a product is detected, the server saves the product's classification for the current frame and all frames received in the next 0.5 seconds, i.e., the approximate time frame for a product to be inserted into the cart.
The model's prediction for the given product is obtained by aggregating these classifications, and the final prediction is sent to the customer display.
To avoid unnecessary computation, the server closes the process once a new product has been identified for one second.
The client-server architecture design is presented in Figure~\ref{shopping_cart_architecture}.

% \begin{figure}[h]
% \includegraphics[width=0.7\linewidth]{Images/Supplementary_materials/cart_for_supplementary.png}
% \centering
% \caption{The smart shopping cart measurements.}
% \label{smart_shopping_cart_measurements}
% \end{figure}

% \begin{figure}[h]
% \includegraphics[width=1.0\linewidth]{Images/Supplementary_materials/smart_shopping_cart_setup.png}
% \centering
% \caption{The smart shopping cart stand characteristics.}
% \label{smart_shopping_cart_stand}
% \end{figure}

% To enable real-time detection, we implemented a client-server architecture that performs the detection on a remote GPU and present the results locally. 
% The client-server architecture design is presented in Figure~\ref{shopping_cart_architecture}.

% \begin{figure}[h]
% \includegraphics[width=1.0\linewidth]{Images/Supplementary_materials/cart_stans_measurments.png}
% \centering
% \caption{The smart shopping cart measurements.}
% \label{shopping_cart_measurments}
% \end{figure}

\begin{figure}[h]
\includegraphics[width=0.98\linewidth]{Images/Supplementary_materials/smart_shopping_cart_setup1.png}
\centering
\caption{Physical evaluation setup.}
\label{Physical_setup}
\end{figure}

\begin{figure}[h]
\includegraphics[width=1.0\linewidth]{Images/Supplementary_materials/shopping_cart_architecture1.png}
\centering
\caption{The smart shopping cart client-server architecture. Frames captured from the cameras on the client side are sent to an object detection model on the server side. 
The model detects the product inserted into the cart and sends the updated shopping list back to the customer display on the client side.
This architecture enables real-time product detection.}
\label{shopping_cart_architecture}
\end{figure}

\begin{figure*}[b]
\includegraphics[width=0.93\linewidth]{Images/Supplementary_materials/SuperStore_items.png}
\centering
\caption{The classes in the Superstore dataset.}
\label{SuperStore_items}
\end{figure*}

\subsection{Superstore Dataset}

The Superstore dataset is dedicated to the object detection domain in the retail sector and the smart shopping cart use-case.
Each image in the dataset was recorded using the smart shopping cart cameras while an item is inserted into the cart (please see additional explanations in our paper).  
The Superstore dataset contains 2,200 images (1,600 for training and 600 for testing), which are evenly distributed across 20 superstore items (classes).
Each image size is 640*360 and annotated by the following attributes:
\begin{itemize}
    \item Image id - the image identification.
    \item Label – a label id (number) and a label name (string) of the main item in the image.
    \item Bounding box in the YOLO format (x,y,w,h).
    \item Bounding box in the Faster R-CNN format (x1,y1,x2,y2).
    \item Light - whether the image is taken in optimal light conditions.
    \item Expensive - whether the item in the image is expensive.
    \item Hand location - hand location on the item inserted (top/side/bottom).
    \item Background – what is the item's background? (office/products in the background).
    \item Visual angle – The angle of the item inserted into the cart (left/straight/right).
\end{itemize}
Figures~\ref{SuperStore_items}, \ref{Superstore additional information} and \ref{Superstore additional information 2} present additional information about the dataset. 
The complete dataset will be made publicly available when the paper is published.

\subsection{Adversarial Patches Crafting Process}

To assess the effectiveness of \name in the physical space, we applied two attacks resulting in the creation of 17 distinct patches, and examples of these patches are shown in Figure ~\ref{Patches}.
These patches were then printed in various sizes, ranging from 200x200 pixels to 400x400 pixels.
Each patch was utilized to attack our smart shopping cart setup.
Notably, we used the smallest patch that effectively deceived the target model was placed on the expensive item.

\begin{figure*}[t]
    \centering
    \begin{subfigure}{0.25\linewidth}
        \centering
        \includegraphics[width=0.85\linewidth]{Images/Supplementary_materials/expensive.png}
        \caption{"Expensive"}
    \end{subfigure}
    \begin{subfigure}{0.25\linewidth}
        \centering
        \includegraphics[width=0.85\linewidth]{Images/Supplementary_materials/light.png}
        \caption{"Light"}
        \label{subfig:digital_blue}
    \end{subfigure}
    \begin{subfigure}{0.25\linewidth}
        \centering
        \includegraphics[width=0.85\linewidth]{Images/Supplementary_materials/hand_location.png}
        \caption{"hand Location"}
    \end{subfigure}
    \begin{subfigure}{0.25\linewidth}
        \centering
        \includegraphics[width=0.85\linewidth]{Images/Supplementary_materials/background.png}
        \caption{"Background"}
    \end{subfigure}
    \begin{subfigure}{0.25\linewidth}
        \centering
        \includegraphics[width=0.85\linewidth]{Images/Supplementary_materials/visual_angle.png}
        \caption{"Visual Angle"}
    \end{subfigure}
    \caption{Superstore Dataset's Distribution.}
    \label{Superstore additional information}
\end{figure*}

\begin{figure*}[t!]
    \centering
    \begin{subfigure}{0.3\linewidth}
        \centering
        \includegraphics[width=0.97\linewidth]{Images/Supplementary_materials/expensive_example.png}
        % \caption{"Expensive"}
    \end{subfigure}
    \begin{subfigure}{0.3\linewidth}
        \centering
        \includegraphics[width=0.97\linewidth]{Images/Supplementary_materials/light_example.png}
        % \caption{"Light"}
        \label{subfig:digital_blue}
    \end{subfigure}
    \begin{subfigure}{0.3\linewidth}
        \centering
        \includegraphics[width=0.97\linewidth]{Images/Supplementary_materials/background_example.png}
        % \caption{"Background"}
    \end{subfigure}
    \begin{subfigure}{0.4\linewidth}
        \centering
        \includegraphics[width=0.97\linewidth]{Images/Supplementary_materials/hand_location_example.png}
        % \caption{"hand Location"}
    \end{subfigure}
    \begin{subfigure}{0.4\linewidth}
        \centering
        \includegraphics[width=0.97\linewidth]{Images/Supplementary_materials/visual_angle_examples.png}
        % \caption{"Visual Angle"}
    \end{subfigure}
    \caption{Superstore Dataset's examples.}
    \label{Superstore additional information 2}
\end{figure*}

\begin{figure*}[h]
\includegraphics[width=1.00\linewidth]{Images/Paper/patches3.png}
\centering
\caption{Examples of the various patches used in the evaluation in the physical space.}
\label{Patches}
\end{figure*}

\subsection{Existing Methods Implementation}
To properly evaluate \name, we implemented two state-of-the-art detectors, Ad-YOLO and Segment \& Complete (SAC), and compared them to \name (please see additional information in our paper).
In the implementation of Ad-YOLO, the target object detection model was extended with an additional class - the "adversarial" class.
To train Ad-YOLO, we extended the datasets to include samples of the \textit{adversarial patch} class, which consists of scenes with objects on which adversarial patches were placed.
The patches used were crafted for this purpose alone and were not part of the original evaluation sets.
The scenes were manually annotated with a bounding box and a classification and were added to the original datasets.
Then, the target models (described in the Experimental Settings Section) were trained on the expanded datasets.
The models were tested on scenes with objects containing similar patches and achieved 98\% accuracy for the \textit{adversarial patch} class.

In SAC, an additional instance segmentation model (UNet) was trained to detect and remove the adversarial patches.
Although SAC is originally designed to improve the robustness against digital disappearance attacks, it can be adjusted to detect illusion attacks in digital and physical spaces.  
We performed the required adjustment by comparing the target model prediction before and after applying the SAC model, which is supposed to detect and remove the patch.
For the digital evaluation, we used the original patches from the paper, since the digital evaluation was performed on the same dataset (COCO). 
For the physical evaluation, we fine-tuned the original SAC UNet model using patches that were crafted with the DPatch and Lee \& Kotler attacks in diverse sizes and shapes based on our SuperStore dataset. 
By training on existing attacks, the segmentation model is exposed to "adversarial” patterns, which are expected to result in a high detection rate for the patches that the detector was trained on. 
However, when challenged by “new” patches (patches that were crafted by a different attack or using a different dataset), SAC performance is expected to be decreased.
The adjusted SAC detector was tested on scenes with objects containing different patches and achieved satisfying results.

\section{Experimental Results Additional Information}

\subsection{Adaptive Attacks Implementation and Results}
In the context of adversarial machine learning, an adaptive attack is a sophisticated attack that is designed to deceive the target model (in our case mislead the object detector) and evade a specific defense method that is assumed to be known to the attacker (\name in our case). 
This type of attack takes into account the defense parameters during the optimization process, with the goal of making the resulted patch more robust to the defense. 
It is important to note that since in an adaptive attack the attacker adds another component to the optimization process, i.e., the component that should make sure that the attack will not be detected by the known defense mechanism, clearly, such an attack is considered more difficult to perform.
 three main goals of an adaptive attack are: (1) to converge; (2) to deceive the target model (i.e., object detector); and (3) to evade the given defense (i.e., \name).

The first goal refers to the ability of the attack to converge, i.e., the attack ability to succeed in reducing the loss during the optimization process despite the incorporation of additional defense parameters. If the attack did not succeed to converge, it means the employed defense mechanism is effective, and cannot compute a patch that can deceive the target model, and at the same time evade the defense method. 
In other words, the fact that the attack optimization process cannot converge means that the defense parameters are so complex that they limit the optimization process, causing the adaptive attack to fail.

Assuming that the attack did converge, it may fail in achieving the second goal of deceiving the target model since the gradients’ landscape of the optimization process becomes more complex and restricted when the defense parameters are taken into account.

Finally, the third goal is to evade the given defense more effectively than the original attack.

To summarize, when comparing the original attack with the adaptive attack, given an effective defense mechanism, we expect that it will be harder for the adaptive attack to converge, it will be less successful in fooling the target model, however, it will be less detected by the defense mechanism.

In our evaluation, we specifically crafted adaptive attacks that evade X-Detect in a white-box setting, i.e., \textbf{full knowledge} of the parameters of the object extraction and scene processing base detectors. 
That was accomplished by adding two additional components to the adaptive attack loss: the OE\_SIFT component and the P\_ASP component. 
The OE\_SIFT component attempts to make the patch more robust to the prototype-KNN classifier by causing the optimizer to maximize the number of matching points between the attacked item and the target item prototypes. 
The P\_ASP component attempts to ensure that the crafted patch is more robust to the SPD image processing techniques by adding those processing techniques as additional transformations to the expectation over transformation used in the optimization process of the patch.

When we attempted to incorporate the exact defense mechanism knowledge within the attack, i.e., the prototype-KNN classifier and scene processing techniques used by X-Detect, the adaptive attack \textbf{failed} to converge (as shown in Figure ~\ref{adaptive_attack}). 
This demonstrates the effectiveness of our proposed defense mechanism. 
This can be explained by the low number of matching points between the attack and target item prototypes in the areas where the patch did not cover the item. 
In addition, the scene processing techniques used by X-Detect (the level of random noise and blurring) caused a dramatic and contrastive change to the scene.
\begin{figure*}[t]
\includegraphics[width=0.75\linewidth]{Images/Supplementary_materials/adaptive_attack.png}
\centering
\caption{The physical adaptive attack crafting results.}
\label{adaptive_attack}
\end{figure*}

In order to further evaluate the effectiveness of our adversarial detector, \name, we used a more \textbf{relaxed versions} of \name components within the adaptive attack (to allow the attack to converge). 
Specifically, we adjusted the OE\_SIFT component to have less effect on the complete loss by adding log to the prototype-KNN classifier’s output and used the NLLLoss function.  
Additionally, we decreased some of the parameters of the SPD image processing techniques (the blurring from 6 to 1 and the random noise from 0.35 to 0.25), which caused less dramatic alteration to the scene.  
This time those attacks converged successfully, allowing us to print three adaptive patches (one for each sub-detector and one for the ensemble) and record additional 300 evaluation videos using our smart shopping cart setup.

While these adaptive patches partially succeeded in deceiving the target object detector, they \textbf{did not} succeed in achieving the third goal of evading \name and the detection rate of \name when applying these adaptive patches remains the same as in the non-adaptive patches.

In summary, crafting an adversarial patch that satisfies all three goals within the context of the smart shopping cart when \name is applied, is almost impossible.  
We showed that evading \name is not possible even by relaxing some of \name parameters, and without that relaxation, the adaptive attacks did not converge. We would like to note that we provided all the code for reproducing the adaptive attacks and validating our results. 

\subsection{Elaborated Experimental Results}
In this subsection, we provide additional information regarding \name's evaluation in the digital and physical space in all attack scenarios.
Tables 1-4 show the digital space evaluation results, where Table 1 presents the attack results and Tables 2-4 present the detection results on the white-box, model-specific and model-agnostic attack scenarios respectively.
Tables 5-12 show the physical space evaluation results, where Tables 5-8 present the attack results and Tables 9-12 present the detection results on the white-box, gray-box, model-specific and model-agnostic attack scenarios respectively.

\begin{table*}[b]
\centering
\label{Attack result on white-box, model specific and model agnostic attack scenarios on digital use-case}
\caption{Attack results in the digital space.}
\begin{adjustbox}{width=0.5\linewidth,center}
\begin{tabular}{c|c|cc|}
\cline{2-4}
\multicolumn{1}{l|}{} & Benign & \multicolumn{2}{c|}{DPatch} \\ \hline
\multicolumn{1}{|c|}{Attack Scenario} & Map & \multicolumn{1}{c|}{Map} & \% of successful attack scenes \\ \hline
\multicolumn{1}{|c|}{White-box} & 0.65 & \multicolumn{1}{c|}{0.21} & 95\% \\ \hline
\multicolumn{1}{|c|}{Model Specific} & 0.2 & \multicolumn{1}{c|}{0.05} & 35\% \\ \hline
\multicolumn{1}{|c|}{Model Agnostic} & 0.17 & \multicolumn{1}{c|}{0.06} & 52\% \\ \hline
\end{tabular}
\end{adjustbox}

\end{table*}

\begin{table*}[b]
\centering
\label{Detection result on white-box attack scenario on digital use-case}
\caption{Detection result for the white-box attack scenario in the digital space.}
\begin{adjustbox}{width=0.85\linewidth,center}
\begin{tabular}{|c|c|c|c|c|}
\hline
Evaluation Metric & Object Extraction  Detector & Scene Processing Detector & MV Ensemble & 2-Tier Ensemble \\ \hline
DA & 0.9 & \textbf{0.91} & \textbf{0.91} & \textbf{0.91} \\ \hline
TPR & \textbf{0.96} & 0.86 & 0.915 & 0.83 \\ \hline
TNR & 0.84 & 0.95 & 0.9 & \textbf{0.99} \\ \hline
FPR & 0.16 & 0.05 & 0.1 & \textbf{0.01} \\ \hline
FNR & \textbf{0.04} & 0.14 & 0.88 & 0.17 \\ \hline
\end{tabular}
\end{adjustbox}
\end{table*}

\begin{table*}[b]
\centering
\label{Detection result on model specific attack scenario on digital use-case}
\caption{Detection result for the model-specific attack scenario in the digital space.}
\begin{adjustbox}{width=0.85\linewidth,center}
\begin{tabular}{|c|c|c|c|c|c|}
\hline
Architecture & Evaluation Metric & Object Extraction Detector & Scene Processing Detector & MV Ensemble & 2-Tier Ensemble \\ \hline
\multirow{4}{*}{\begin{tabular}[c]{@{}c@{}}Faster R-CNN \\ ResNet-50-FPN\end{tabular}} & DA & \textbf{0.93} & 0.855 & 0.87 & 0.89 \\ \cline{2-6} 
 & TPR & \textbf{1} & 0.83 & 0.875 & 0.83 \\ \cline{2-6} 
 & TNR & 0.86 & 0.88 & 0.87 & \textbf{0.95} \\ \cline{2-6} 
& FPR & 0.14 & 0.12 & 0.13 & \textbf{0.05} \\ \cline{2-6} 
& FNR & \textbf{0.0} & 0.17 & 0.125 & 0.17 \\ \hline 
\multirow{4}{*}{\begin{tabular}[c]{@{}c@{}}Faster R-CNN \\ ResNet-50-FPN \\ IOU Loss\end{tabular}} & DA & 0.895 & 0.905 & 0.89 & \textbf{0.92} \\ \cline{2-6} 
 & TPR & \textbf{0.95} & 0.93 & 0.919 & 0.88 \\ \cline{2-6} 
 & TNR & 0.84 & 0.88 & 0.86 & \textbf{0.96} \\ \cline{2-6} 
 & FPR & 0.16 & 0.12 & 0.14 & \textbf{0.04} \\ \cline{2-6} 
& FNR & \textbf{0.05} & 0.07 & 0.081 & 0.12 \\ \hline
\multirow{4}{*}{\begin{tabular}[c]{@{}c@{}}Faster R-CNN \\ ResNet-101-FPN\end{tabular}} & DA & 0.825 & 0.855 & \textbf{0.899} & 0.818 \\ \cline{2-6} 
 & TPR & 0.83 & 0.77 & \textbf{0.888} & 0.66 \\ \cline{2-6} 
 & TNR & 0.82 & 0.94 & 0.9 & \textbf{0.976} \\ \cline{2-6} 
 & FPR & 0.18 & 0.06 & 0.1 & \textbf{0.024} \\ \cline{2-6} 
& FNR & 0.17 & 0.23 & \textbf{0.11} & 0.34 \\ \hline

\end{tabular}
\end{adjustbox}
\end{table*}

\begin{table*}[b]
\centering
\label{Detection result on model agnostic attack scenario on digital use-case}
\caption{Detection result for the model agnostic attack scenario in the digital space.}
\begin{adjustbox}{width=0.85\linewidth,center}
\begin{tabular}{|c|c|c|c|c|c|}
\hline
Architecture & Evaluation Metric & Object Extraction Detector & Scene Processing Detector & MV Ensemble & 2-Tier Ensemble \\ \hline
\multirow{4}{*}{Grid RCNN} & DA & 0.86 & 0.795 & \textbf{0.865} & 0.855 \\ \cline{2-6} 
 & TPR & \textbf{0.93} & 0.81 & \textbf{0.93} & 0.79 \\ \cline{2-6} 
 & TNR & 0.79 & 0.78 & 0.8 & \textbf{0.92} \\ \cline{2-6} 
 & FPR & 0.21 & 0.28 & 0.2 & \textbf{0.08} \\ \cline{2-6} 
& FNR & \textbf{0.07} & 0.19 & \textbf{0.07} & 0.21 \\ \hline
\multirow{4}{*}{Cascade RCNN} & DA & \textbf{0.895} & 0.87 & 0.875 & \textbf{0.895} \\ \cline{2-6} 
 & TPR & \textbf{0.94} & 0.83 & 0.87 & 0.8 \\ \cline{2-6} 
 & TNR & 0.85 & 0.915 & 0.88 & \textbf{0.987} \\ \cline{2-6} 
 & FPR & 0.15 & 0.08 & 0.12 & \textbf{0.013} \\ \cline{2-6} 
& FNR & \textbf{0.06} & 0.17 & 0.13 & 0.2 \\ \hline
\end{tabular}
\end{adjustbox}
\end{table*}

\begin{table*}[b]
\centering
\label{Attack result on white-box attack scenario}
\caption{Attack result for the white-box attack scenario in the physical space.}
\begin{adjustbox}{width=0.85\linewidth,center}
\begin{tabular}{cc|ccc|ccc|}
\cline{3-8}
\multicolumn{1}{l}{} & \multicolumn{1}{l|}{} & \multicolumn{3}{c|}{Shopping list level} & \multicolumn{3}{c|}{Scene level} \\ \cline{2-8} 
\multicolumn{1}{l|}{} & Attack
 
& \multicolumn{1}{c|}{\begin{tabular}[c]{@{}c@{}}\% of target  \\ item in the list\end{tabular}} 
& \multicolumn{1}{c|}{\begin{tabular}[c]{@{}c@{}}\% of cheap  \\ item in the list\end{tabular}} 
& \multicolumn{1}{c|}{\begin{tabular}[c]{@{}c@{}}\% of wrong  \\ item in the list\end{tabular}} 
& \multicolumn{1}{c|}{\begin{tabular}[c]{@{}c@{}}\% of target  \\ class predicted\end{tabular}} 
& \multicolumn{1}{c|}{\begin{tabular}[c]{@{}c@{}}\% of cheap  \\ class predicted\end{tabular}} 
& \multicolumn{1}{c|}{\begin{tabular}[c]{@{}c@{}}\% of wrong  \\ class predicted\end{tabular}} \\ \hline
\multicolumn{1}{|c|}{\multirow{3}{*}{\begin{tabular}[c]{@{}c@{}}Faster R-CNN\\ Resnet-50-FPN \\ Seed 42\end{tabular}}} & Dpatch & \multicolumn{1}{c|}{0.80} & \multicolumn{1}{c|}{0.98} & 0.98 & \multicolumn{1}{c|}{0.61} & \multicolumn{1}{c|}{0.79} & 0.91 \\ \cline{2-8} 
\multicolumn{1}{|c|}{} & Lee \& Kotler & \multicolumn{1}{c|}{0.80} & \multicolumn{1}{c|}{0.99} & 1 & \multicolumn{1}{c|}{0.61} & \multicolumn{1}{c|}{0.81} & 0.93 \\ \cline{2-8} 
\multicolumn{1}{|c|}{} & Overall & \multicolumn{1}{c|}{\textbf{0.8 (1120)}} & \multicolumn{1}{c|}{0.98} & 0.99 & \multicolumn{1}{c|}{0.61} & \multicolumn{1}{c|}{0.8} & 0.92 \\ \hline
\end{tabular}
\end{adjustbox}
\end{table*}

\begin{table*}[b]
\centering
\label{Attack result on grey-box attack scenario}
\caption{Attack result for grey-box attack scenario in the physical space.}
\begin{adjustbox}{width=0.85\linewidth,center}
\begin{tabular}{cc|cc|cc|}
\cline{3-6}
\multicolumn{1}{l}{} & \multicolumn{1}{l|}{} & \multicolumn{2}{c|}{Shopping list level} & \multicolumn{2}{c|}{Scene level} \\ \hline
\multicolumn{1}{|c|}{Seed} & Attack & \multicolumn{1}{c|}{\% of cheap item in the list} & \% of wrong item in the list & \multicolumn{1}{c|}{\% of cheap class predicted} & \% of wrong class predicted \\ \hline
\multicolumn{1}{|c|}{\multirow{3}{*}{38}} & Dpatch & \multicolumn{1}{c|}{0.62} & 0.86 & \multicolumn{1}{c|}{0.6} & 0.86 \\ \cline{2-6} 
\multicolumn{1}{|c|}{} & Lee \& Kotler & \multicolumn{1}{c|}{0.59} & 0.91 & \multicolumn{1}{c|}{0.57} & 0.91 \\ \cline{2-6} 
\multicolumn{1}{|c|}{} & Overall & \multicolumn{1}{c|}{\textbf{0.6 (840)}} & 0.89 & \multicolumn{1}{c|}{0.58} & 0.89 \\ \hline
\multicolumn{1}{|c|}{\multirow{3}{*}{40}} & Dpatch & \multicolumn{1}{c|}{0.63} & 0.82 & \multicolumn{1}{c|}{0.6} & 0.81 \\ \cline{2-6} 
\multicolumn{1}{|c|}{} & Lee \& Kotler & \multicolumn{1}{c|}{0.58} & 0.86 & \multicolumn{1}{c|}{0.55} & 0.84 \\ \cline{2-6} 
\multicolumn{1}{|c|}{} & Overall & \multicolumn{1}{c|}{\textbf{0.6 (840)}} & 0.84 & \multicolumn{1}{c|}{0.57} & 0.83 \\ \hline
\multicolumn{1}{|c|}{\multirow{3}{*}{44}} & Dpatch & \multicolumn{1}{c|}{0.75} & 0.85 & \multicolumn{1}{c|}{0.73} & 0.84 \\ \cline{2-6} 
\multicolumn{1}{|c|}{} & Lee \& Kotler & \multicolumn{1}{c|}{0.74} & 0.9 & \multicolumn{1}{c|}{0.75} & 0.88 \\ \cline{2-6} 
\multicolumn{1}{|c|}{} & Overall & \multicolumn{1}{c|}{\textbf{0.75 (1050)}} & 0.88 & \multicolumn{1}{c|}{0.72} & 0.87 \\ \hline
\end{tabular}
\end{adjustbox}
\end{table*}

\begin{table*}[b]
\centering
\label{Attack result on model specific attack scenario}
\caption{Attack result for the model specific attack scenario in the physical space.}
\begin{adjustbox}{width=0.85\linewidth,center}
\begin{tabular}{lc|cc|cc|}
\cline{3-6}
 & \multicolumn{1}{l|}{} & \multicolumn{2}{c|}{Shopping list level} & \multicolumn{2}{c|}{Scene   level} \\ \hline
\multicolumn{1}{|l|}{Seed} & \multicolumn{1}{l|}{Attack} & \multicolumn{1}{c|}{\% of cheap item in the list} & \% of wrong item in the list & \multicolumn{1}{c|}{\% of cheap class predicted} & \% of wrong class predicted \\ \hline
\multicolumn{1}{|l|}{\multirow{3}{*}{\begin{tabular}[c]{@{}l@{}}Faster R-CNN\\ ResNet-50-FPN\end{tabular}}} & Dpatch & \multicolumn{1}{c|}{0.44} & 0.52 & \multicolumn{1}{c|}{0.43} & 0.53 \\ \cline{2-6} 
\multicolumn{1}{|l|}{} & Lee \& Kotler & \multicolumn{1}{c|}{0.18} & 0.49 & \multicolumn{1}{c|}{0.17} & 0.5 \\ \cline{2-6} 
\multicolumn{1}{|l|}{} & Overall & \multicolumn{1}{c|}{\textbf{0.29 (406)}} & 0.51 & \multicolumn{1}{c|}{0.28} & 0.51 \\ \hline
\multicolumn{1}{|l|}{\multirow{3}{*}{\begin{tabular}[c]{@{}l@{}}Faster R-CNN \\ ResNet-50-FPN\\ IOU Loss\end{tabular}}} & Dpatch & \multicolumn{1}{c|}{0.51} & 0.57 & \multicolumn{1}{c|}{0.51} & 0.59 \\ \cline{2-6} 
\multicolumn{1}{|l|}{} & Lee \& Kotler & \multicolumn{1}{c|}{0.26} & 0.56 & \multicolumn{1}{c|}{0.26} & 0.56 \\ \cline{2-6} 
\multicolumn{1}{|l|}{} & Overall & \multicolumn{1}{c|}{\textbf{0.36 (504)}} & 0.56 & \multicolumn{1}{c|}{0.36} & 0.57 \\ \hline
\multicolumn{1}{|l|}{\multirow{3}{*}{\begin{tabular}[c]{@{}l@{}}Faster R-CNN \\ ResNet-101-FPN\end{tabular}}} & Dpatch & \multicolumn{1}{c|}{0.35} & 0.46 & \multicolumn{1}{c|}{0.37} & 0.48 \\ \cline{2-6} 
\multicolumn{1}{|l|}{} & Lee \& Kotler & \multicolumn{1}{c|}{0.25} & 0.57 & \multicolumn{1}{c|}{0.26} & 0.58 \\ \cline{2-6} 
\multicolumn{1}{|l|}{} & Overall & \multicolumn{1}{c|}{\textbf{0.29 (406)}} & 0.52 & \multicolumn{1}{c|}{0.31} & 0.54 \\ \hline
\end{tabular}
\end{adjustbox}
\end{table*}

\begin{table*}[b]
\centering
\label{Attack result on model agnostic attack scenario}
\caption{Attack result for the model agnostic attack scenario in the physical space.}
\begin{adjustbox}{width=0.85\linewidth,center}
\begin{tabular}{cc|cc|cc|}
\cline{3-6}
\multicolumn{1}{l}{} & \multicolumn{1}{l|}{} & \multicolumn{2}{c|}{Shopping list level} & \multicolumn{2}{c|}{Scene level} \\ \hline
\multicolumn{1}{|c|}{OD Algorithm} & Attack & \multicolumn{1}{c|}{\% of cheap item in the list} & \% of wrong item in the list & \multicolumn{1}{c|}{\% of cheap class predicted} & \% of wrong class predicted \\ \hline \hline
\multicolumn{1}{|c|}{\multirow{3}{*}{\begin{tabular}[c]{@{}c@{}}Cascade \\ R-CNN\end{tabular}}} & Dpatch & \multicolumn{1}{c|}{0.39} & 0.57 & \multicolumn{1}{c|}{0.37} & 0.54 \\ \cline{2-6} 
\multicolumn{1}{|c|}{} & Lee \& Kotler & \multicolumn{1}{c|}{0.24} & 0.57 & \multicolumn{1}{c|}{0.24} & 0.56 \\ \cline{2-6} 
\multicolumn{1}{|c|}{} & Overall & \multicolumn{1}{c|}{\textbf{0.31 (434)}} & 0.57 & \multicolumn{1}{c|}{0.3} & 0.55 \\ \hline 
\multicolumn{1}{|c|}{\multirow{3}{*}{\begin{tabular}[c]{@{}c@{}}Cascade \\ RPN\end{tabular}}} & Dpatch & \multicolumn{1}{c|}{0.52} & 0.65 & \multicolumn{1}{c|}{0.51} & 0.64 \\ \cline{2-6} 
\multicolumn{1}{|c|}{} & Lee \& Kotler & \multicolumn{1}{c|}{0.34} & 0.67 & \multicolumn{1}{c|}{0.33} & 0.65 \\ \cline{2-6} 
\multicolumn{1}{|c|}{} & Overall & \multicolumn{1}{c|}{\textbf{0.42 (588)}} & 0.66 & \multicolumn{1}{c|}{0.41} & 0.65 \\ \hline 
\multicolumn{1}{|c|}{\multirow{3}{*}{YOLOv3}} & Dpatch & \multicolumn{1}{c|}{0.12} & 0.69 & \multicolumn{1}{c|}{0.1} & 0.63 \\ \cline{2-6} 
\multicolumn{1}{|c|}{} & Lee \& Kotler & \multicolumn{1}{c|}{0.12} & 0.66 & \multicolumn{1}{c|}{0.11} & 0.65 \\ \cline{2-6} 
\multicolumn{1}{|c|}{} & Overall & \multicolumn{1}{c|}{\textbf{0.12 (168)}} & 0.67 & \multicolumn{1}{c|}{0.1} & 0.64 \\ \hline
\end{tabular}
\end{adjustbox}
\end{table*}

\begin{table*}[b]
\centering
\label{Detection results on white-box attack scenario on the physical use-case}
\caption{Detection results for the white-box attack scenario in the physical space.}
\begin{adjustbox}{width=0.85\linewidth,center}
\begin{tabular}{|c|c|c|c|c|c|}
\hline
\multicolumn{1}{|l|}{Evaluation Metric}  & \multicolumn{1}{l|}{Object Extraction Detector} & \multicolumn{1}{l|}{Scene Processing Detector} & \multicolumn{1}{l|}{MV Ensemble} & \multicolumn{1}{l|}{2-Tier Ensemble} \\ \hline
DA & 0.84 & 0.913 & \textbf{0.96} & 0.939 \\ \hline
TPR & \textbf{0.98} & 0.894 & 0.92 & 0.879 \\ \hline
TNR & 0.7 & 0.933 & \textbf{1} & \textbf{1} \\ \hline
FNR & 0.3 & 0.067 & \textbf{0} & \textbf{0} \\ \hline
FPR & \textbf{0.02} & 0.106 & 0.08 & 0.121 \\ \hline
\end{tabular}
\end{adjustbox}
\end{table*}

\begin{table*}[b]
\label{Detection results on grey-box attack scenario on the physical use-case}
\caption{Detection results for the grey-box attack scenario in the physical space.}
\begin{adjustbox}{width=0.85\linewidth,center}
\begin{tabular}{|c|c|c|c|c|c|}
\hline
Seed & Evaluation Metric & Object Extraction Detector & Scene Processing Detector & MV Ensemble & 2-Tier Ensemble \\ \hline
\multirow{5}{*}{38} & DA & 0.84 & 0.836 & \textbf{0.926} & 0.875 \\ \cline{2-6} 
 & TPR & \textbf{0.98} & 0.74 & 0.853 & 0.75 \\ \cline{2-6} 
 & TNR & 0.7 & 0.933 & \textbf{1} & \textbf{1} \\ \cline{2-6} 
 & FNR & 0.3 & 0.067 & \textbf{0} & \textbf{0} \\ \cline{2-6} 
 & FPR & \textbf{0.02} & 0.26 & 0.14 & 0.25 \\ \hline
\multirow{5}{*}{40} & DA & 0.84 & 0.83 & \textbf{0.914} & 0.88 \\ \cline{2-6} 
 & TPR & \textbf{0.98} & 0.723 & 0.828 & 0.76 \\ \cline{2-6} 
 & TNR & 0.7 & 0.933 & \textbf{1} & \textbf{1} \\ \cline{2-6} 
 & FNR & 0.3 & 0.067 & \textbf{0} & \textbf{0} \\ \cline{2-6} 
 & FPR & \textbf{0.02} & 0.27 & 0.17& 0.24 \\ \hline
\multirow{5}{*}{44} & DA & 0.83 & 0.86 & \textbf{0.895} & 0.87 \\ \cline{2-6} 
 & TPR & \textbf{0.97} & 0.731 & 0.788 & 0.736 \\ \cline{2-6} 
 & TNR & 0.7 & 0.933 & \textbf{1} & \textbf{1} \\ \cline{2-6} 
 & FNR & 0.3 & 0.067 & \textbf{0} & \textbf{0} \\ \cline{2-6} 
 & FPR & \textbf{0.03} & 0.269 & 0.21 & 0.264 \\ \hline
\end{tabular}
\end{adjustbox}
\end{table*}

\begin{table*}[b]
\label{Detection results on model specific attack scenario on the physical use-case}
\caption{Detection results for the model specific attack scenario in the physical space.}
\begin{adjustbox}{width=0.85\linewidth,center}
\begin{tabular}{|c|c|c|c|c|c|}
\hline
Architecture & Evaluation Metric & Object Extraction Detector & Scene Processing Detector & MV Ensemble & 2-Tier Ensemble \\ \hline
\multirow{5}{*}{\begin{tabular}[c]{@{}c@{}}Faster RCNN \\ ResNet-50-FPN\end{tabular}} & DA & 0.83 & 0.82 & \textbf{0.95} & 0.91 \\ \cline{2-6} 
 & TPR & \textbf{0.96} & 0.7 & 0.9 & 0.82 \\ \cline{2-6} 
 & TNR & 0.7 & 0.933 & \textbf{1} & \textbf{1} \\ \cline{2-6} 
 & FNR & 0.3 & 0.067 & \textbf{0} & \textbf{0} \\ \cline{2-6} 
 & FPR & \textbf{0.04} & 0.3 & 0.1 & 0.18 \\ \hline
\multirow{5}{*}{\begin{tabular}[c]{@{}c@{}}Faster RCNN \\ ResNet-50-FPN\\ IOU Loss\end{tabular}} & DA & 0.82 & 0.82 & \textbf{0.895} & 0.835 \\ \cline{2-6} 
 & TPR & \textbf{0.94} & 0.7 & 0.79 & 0.67 \\ \cline{2-6} 
 & TNR & 0.7 & 0.933 & \textbf{1} & \textbf{1} \\ \cline{2-6} 
 & FNR & 0.3 & 0.067 & \textbf{0} & \textbf{0} \\ \cline{2-6} 
 & FPR & \textbf{0.06} & 0.3 & 0.21 & 0.33 \\ \hline
\multirow{5}{*}{\begin{tabular}[c]{@{}c@{}}Faster RCNN \\ ResNet-101-FPN\end{tabular}} & DA & 0.82 & 0.89 & \textbf{0.963} & 0.945 \\ \cline{2-6} 
 & TPR & \textbf{0.948} & 0.85 & 0.927 & 0.89 \\ \cline{2-6} 
 & TNR & 0.7 & 0.933 & \textbf{1} & \textbf{1} \\ \cline{2-6} 
 & FNR & 0.3 & 0.067 & \textbf{0} & \textbf{0} \\ \cline{2-6} 
 & FPR & \textbf{0.052} & 0.15 & 0.073 & 0.11 \\ \hline
\end{tabular}
\end{adjustbox}
\end{table*}

\begin{table*}[b]
\label{Detection results on model agnostic attack scenario on the physical use-case}
\caption{Detection results for the model agnostic attack scenario in the physical space.}
\begin{adjustbox}{width=0.85\linewidth,center}
\begin{tabular}{|c|c|c|c|c|c|}
\hline
Architecture & Evaluation Metric & Object Extraction Detector & Scene Processing Detector & MV Ensemble & 2-Tier Ensemble \\ \hline
\multirow{5}{*}{\begin{tabular}[c]{@{}c@{}}Cascade \\ R-CNN\end{tabular}} & DA & 0.81 & 0.78 & \textbf{0.86} & 0.83 \\ \cline{2-6} 
 & TPR & \textbf{0.94} & 0.71 & \textbf{0.81} & 0.70 \\ \cline{2-6} 
 & TNR & 0.69 & 0.85 & 0.92 & \textbf{0.96} \\ \cline{2-6} 
 & FNR & 0.31 & 0.15 & \textbf{0.08} & \textbf{0.04} \\ \cline{2-6} 
 & FPR & \textbf{0.06} & 0.29 & 0.19 & 0.3 \\ \hline
\multirow{5}{*}{\begin{tabular}[c]{@{}c@{}}Cascade \\ RPN\end{tabular}} & DA & \textbf{0.81} & 0.89 & 0.94 & \textbf{0.89} \\ \cline{2-6} 
 & TPR & \textbf{0.94} & 0.83 & 0.89 & 0.80 \\ \cline{2-6} 
 & TNR & 0.69 & 0.96 & \textbf{0.98} & \textbf{0.98} \\ \cline{2-6} 
 & FNR & 0.31 & 0.04 & \textbf{0.02} & \textbf{0.02} \\ \cline{2-6} 
 & FPR & \textbf{0.06} & 0.17 & 0.11 & 0.2 \\ \hline
\multirow{5}{*}{YOLOv3} & DA & 0.82 & 0.86 & \textbf{0.91} & 0.86 \\ \cline{2-6} 
 & TPR & \textbf{0.96} & 0.77 & 0.83 & 0.74 \\ \cline{2-6} 
 & TNR & 0.69 & 0.96 & \textbf{0.98} & \textbf{0.98} \\ \cline{2-6} 
 & FNR & 0.31 & 0.04 & \textbf{0.02} & \textbf{0.02} \\ \cline{2-6} 
 & FPR & \textbf{0.04} & 0.23 & 0.17 & 0.26 \\ \hline
\end{tabular}
\end{adjustbox}
\end{table*}

%%%%%%%%%%%%%%%%%%%%%%%%%%%%%%%%%%%%%%%%%%%%%%%%%%%%%%%%%%%%